\documentclass[lettersize,journal]{IEEEtran}
\usepackage{amsmath,amsfonts}
\usepackage{algpseudocode}
\usepackage[ruled,vlined,linesnumbered]{algorithm2e}
\usepackage{array}
\usepackage[caption=false,font=normalsize,labelfont=sf,textfont=sf]{subfig}
\usepackage{textcomp}
\usepackage{stfloats}
\usepackage{url}
\usepackage{verbatim}
\usepackage{graphicx}
\usepackage{cite}
\usepackage[dvipsnames]{xcolor}
\usepackage{caption}
\usepackage{multirow}
\usepackage{colortbl}
\usepackage{xcolor}
\usepackage{tikz,lipsum,lmodern}
\usepackage[most]{tcolorbox}
\usepackage[table]{xcolor}
\usepackage{booktabs}
\usepackage{amssymb}
\usepackage{pifont}
\usepackage{mdframed}
\usepackage{orcidlink}
\usepackage[dvipsnames]{xcolor}

\begin{document}

\title{From Generator to Embedder: Harnessing Innate Abilities of Multimodal LLMs via Building Zero-Shot Discriminative Embedding Model}

\author{Yeong-Joon Ju~\orcidlink{0009-0009-5552-2520},
Seong-Whan Lee~\orcidlink{0000-0002-6249-4996}, \IEEEmembership{Fellow,~IEEE},
\thanks{Y.-J. Ju and S.-W. Lee are in the Department of Artificial Intelligence at Korea University, Seoul, Republic of Korea. This work was supported by the Institute of Information \& Communications Technology Planning \& Evaluation (IITP) grants, funded by the Korea government (MSIT) (No. RS-2019-II190079 (Artificial Intelligence Graduate School Program (Korea University)), No. IITP-2025-RS-2024-00436857 (Information Technology Research Center (ITRC)), No. IITP-2026-RS-2025-02304828 (Artificial Intelligence Star Fellowship Support Program to Nurture the Best Talents)). Corresponding author: Seong-Whan Lee. E-mail: yj\_ju@korea.ac.kr, sw.lee@korea.ac.kr}}



\maketitle

\begin{abstract}
Adapting generative Multimodal Large Language Models (MLLMs) into universal embedding models typically demands resource-intensive contrastive pre-training, while traditional hard negative mining methods suffer from severe false negative contamination. In this paper, we propose a highly data-efficient framework that bypasses extensive pre-training to build a robust multimodal representation space. We first introduce a hierarchical embedding prompt that provides strong latent conditioning. By explicitly anchoring task definitions at the system level, this prompting strategy effectively bridges the modality gap and unlocks powerful zero-shot embedding capabilities. Building upon this latent conditioning, we present Self-aware Hard Negative Sampling (SaHa). Unlike conventional candidate-space mining, SaHa shifts the mechanism to the query-space by mapping retrieved candidates back to their owner queries to rigorously filter out semantic false negatives. Furthermore, our method constructs mutually hard clusters, maximizing intra-task discrimination and batch efficiency without redundant forward passes. Extensive experiments demonstrate that our unified approach achieves highly competitive fine-tuning performance on the Massive Multimodal Embedding Benchmark using only a fraction of standard training data. Our code: \url{https://github.com/yeongjoonJu/Gen2Embed}
\end{abstract}


\section{Introduction}

The exponential growth of multimedia applications has fundamentally shifted the research focus from task-specific models toward universal multimodal embeddings~\cite{liu2023universal,zhou-etal-2024-vista,jiang2024e5,jiang2025vlmvec} capable of supporting both uni- and cross-modal retrieval~\cite{wei2025dynamic,pu2025deep,zhuang2023towards,zhang2023weighted}, designed to provide robust representations for diverse downstream tasks such as clustering, retrieval~\cite{karpukhin-etal-2020-dense,lewis2020retrieval}, and classification~\cite{muennighoff-etal-2023-mteb,wang2025visual} within a single framework. While prevailing approaches~\cite{liu2023universal,baldrati2023composed,wei2024uniir} have predominantly relied on dual-encoder architectures like CLIP~\cite{radford2021learning} or BLIP~\cite{li2022blip}, these models often struggle with complex interleaved or text-heavy inputs due to their lack of in-depth image-text fusion~\cite{lin2023fine,luo-etal-2023-end,ju-etal-2025-mire}. To overcome these architectural limitations, Multimodal Large Language Models (MLLMs)~\cite{song2025bridge} have emerged as a promising solution. By leveraging their generative pre-training on vast interleaved data and inheriting powerful instruction-following abilities~\cite{wei2022emergent}, MLLMs are uniquely positioned to capture intricate cross-modal dependencies, offering a compelling foundation for universal embedding tasks.

\begin{figure}
\centering
\includegraphics[width = 1.\columnwidth]{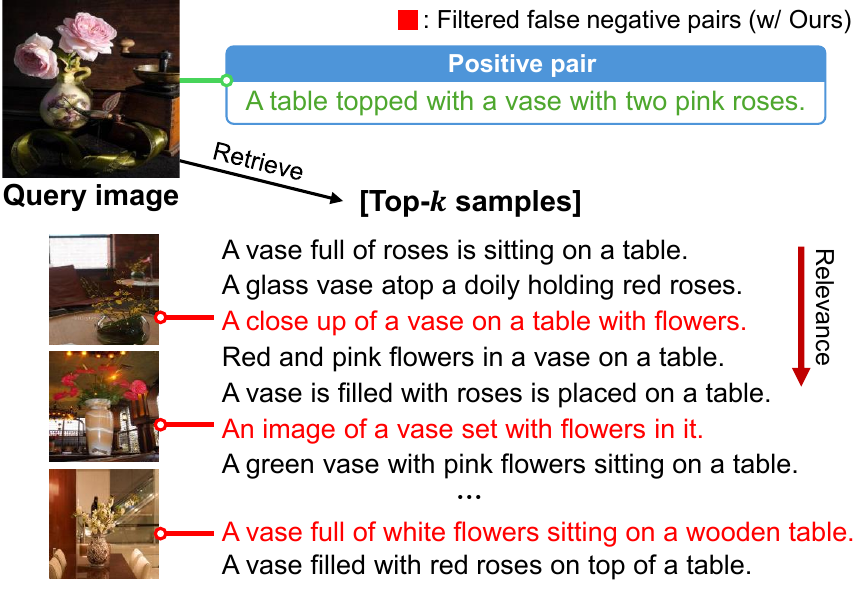}
\captionof{figure}{
    \textbf{False negatives in hard negative mining.} Since discriminative training relies on annotated pairs, valid descriptions \textcolor{red}{(red)} are treated as negatives simply since they are not explicitly paired with the query \textcolor{YellowGreen}{(green)}.}
\vspace{-6pt}
\label{fig:motiv}
\end{figure}


Nevertheless, establishing a robust universal embedding space requires achieving inter-task distinction and intra-task discrimination simultaneously. Recent text-based methods such as SFR-Embedding~\cite{meng2024sfr} and NV-Embed~\cite{lee2025nvembed} have focused on optimizing batch construction, employing task-homogeneous or well-blended batching to mitigate gradient conflicts between diverse tasks. Although these strategies stabilize inter-task learning, they typically rely on standard in-batch negatives, which often provide insufficient discriminative signals for capturing fine-grained intra-task distinctions. To refine the embedding space further, incorporating Hard Negative Sampling (HNS) is essential. Recent approaches~\cite{gu2025breaking,kong2025modality,lan2025llave} leverage external teacher models to learn more discriminative representations by forcing the model to focus on hard negative pairs. However, vanilla hard negative mining inevitably suffers from the false negative issue, where valid positives are erroneously selected as negatives. Such false negatives are prevalent in multimodal retrieval datasets, where semantically equivalent candidates are treated as negatives simply because they lack explicit pairing with the anchor, as shown in Fig.~\ref{fig:motiv}. To alleviate this issue, UniME~\cite{gu2025breaking} attempts to filter out false negatives by applying similarity-based thresholding derived from an external text-based teacher model. Yet, such heuristic thresholds often fail to adapt to the varying similarity distributions across diverse tasks, leading to potentially contradictory signals or the exclusion of informative hard negatives. Furthermore, dependence on external teacher models, typically text-centric retrievers, forces the multimodal student to mimic a unimodal teacher, potentially undermining the MLLM's critical cross-modal alignment. Consequently, there is a distinct lack of a generalized method that is batch-efficient across diverse modalities and capable of leveraging the MLLM's intrinsic multimodal knowledge to resolve these ambiguities without external dependencies.

To bridge this gap, we propose Self-aware Hard negative sampling (SaHa), which harnesses this intrinsic knowledge to distinguish informative hard negatives from conflicting false negatives. Instead of treating the batch as a mere collection of random samples, SaHa exploits the semantic relationships between queries. Our core intuition is that semantically similar queries share similar targets. The positive candidate of one query inherently acts as a false negative for its semantic neighbors. Based on this intuition, SaHa first aggregates a pool of semantically related candidates and identifies their originating owner queries for a given anchor. By strategically selecting candidates whose owner queries are semantically distinct from the anchor, we ensure that the selected negatives are hard yet valid. Crucially, our method constructs a cluster of mutually hard samples, where each candidate acts as both a positive for its owner and a rigorous hard negative for other anchors. This structure maximizes the discriminative signal density within the batch, achieving high training efficiency and robust intra-task discrimination without external dependencies.

Implementing SaHa directly on off-the-shelf MLLMs, however, is non-trivial due to a fundamental functional misalignment. The inherent objective of next-token prediction is optimized for generating coherent sequences, not for producing a single, compact embedding that encapsulates the entire input. To bridge this gap, the dominant paradigm~\cite{jiang2024e5,gu2025breaking,zhang2025bridging,yu2025cafe,ouali2025vladva,kong2025modality} has relied on large-scale contrastive pre-training. While effective, this approach incurs prohibitive computational costs and often requires complex joint training with generation loss to prevent catastrophic forgetting~\cite{yu2025cafe, ouali2025vladva}. Recent studies~\cite{jiang2024e5, ouali2025vladva} have explored a more efficient alternative, demonstrating that appending prompts like ``in one word" can guide MLLMs toward discriminative representations. We advance this direction by investigating the structural impact of different prompt types. Our empirical analysis reveals that system prompts are significantly more effective than user-level instructions for latent conditioning, acting as global anchors. Consequently, they fundamentally constrain the generation space, thereby minimizing the modality gap. Based on this insight, we employ a Hierarchical Embedding Prompt to create a structurally coherent embedding space from the outset. This conditioning serves as the essential foundation that enables SaHa to function effectively from the very first training epoch, bypassing the need for massive pre-training. Our main contributions are summarized as follows:

\begin{itemize}
    \item We propose Self-aware Hard negative sampling (SaHa), a novel mining strategy that exploits the latent semantic structure of the batch to autonomously filter false negatives.

    \item We identify the structural superiority of system-level instructions for latent conditioning and introduce a Hierarchical Embedding Prompt.

    \item We demonstrate the efficacy of our framework through comprehensive experiments. Our method achieves state-of-the-art performance on the massive MMEB benchmark, outperforming existing approaches that rely on large-scale pre-training or external teachers.
\end{itemize}

\section{Related Work}

\subsection{Multimodal Retrieval}
The need to process inputs that combine vision and language has driven the development of multimodal retrieval, moving beyond text retrieval. Early approaches focused on enriching the input by converting images into richer textual representations, such as captions~\cite{gao2022thousand,salemi2023symmetric} or object tags~\cite{gui-etal-2022-kat,yang2022empirical,wu-mooney-2022-entity}.  To eliminate the dependency on such detached modules, subsequent research~\cite{luo-etal-2023-end,ju-etal-2025-mire,zhou-etal-2024-vista} shifted towards end-to-end retrieval that employs in-depth fusion between modalities. Following the success of multi-task training in text embeddings~\cite{su-etal-2023-one,asai-etal-2023-task}, recent studies~\cite{wei2024uniir,zhou-etal-2024-vista} leveraged instruction-following tuning to handle heterogeneous queries and candidates in diverse modalities, pushing the boundaries towards universal multimodal embeddings. However, their architectures with separate initial encoders still posed a challenge in achieving a universal multimodal embedding space.

\subsection{Universal Multimodal Embeddings}

To overcome the task-specific nature of earlier models, research on universal embeddings aims to create a single model that provides robust representations across a diverse range of tasks and domains. The success of LLMs as backbones for text-only embeddings has been demonstrated by several studies~\cite{ma2024fine, li2024llama2vec, lee2025nvembed}, showing that large-scale models possess strong capabilities for producing powerful representations.

While previous approaches~\cite{wei2024uniir,zhou-etal-2024-vista} for universal multimodal embeddings enhanced dual-encoders with more sophisticated fusion, they still operated on separately encoded modalities. This fundamental architectural limitation motivated the shift toward MLLMs. By pre-training on interleaved visual and textual data, MLLMs can capture complex cross-modal dependencies natively, emerging as a promising solution to the weaknesses of dual-encoders. The dominant paradigm for adapting their generative nature for discriminative embedding tasks is large-scale contrastive fine-tuning. VLM2Vec~\cite{jiang2025vlmvec} and MM-Embed~\cite{lin2025mmembed} fine-tune MLLMs on extensive, diverse multimodal task datasets to create a universal embedding space. Others focus on the critical role of training data. MegaPairs~\cite{zhou2024megapairs} and GME~\cite{zhang2025bridging} introduce large-scale data synthesis pipelines to generate high-quality image-text pairs with open-ended instructions, thereby enhancing model performance on universal retrieval benchmarks.

A parallel line of research investigates how to preserve the generative strengths of MLLMs during this discriminative fine-tuning. VladVA~\cite{ouali2025vladva} and CAFe~\cite{yu2025cafe} propose joint training frameworks that combine contrastive loss with a next-token prediction loss. This allows the model to learn discriminative representations without suffering from catastrophic forgetting of its generative capabilities. UNITE~\cite{kong2025modality} has pushed this further by analyzing modality-specific data properties and proposing a modality-aware contrastive learning objective to mitigate interference between different modalities during training. Despite these advances, the prevailing methods still rely on computationally expensive, large-scale contrastive pre-training and complex training objectives. This highlights a clear gap for our efficient framework that unlocks the inherent capabilities of MLLMs.

\section{Method}
\label{sec:method}

In this section, we present an integrated fine-tuning strategy designed to establish a robust universal embedding space. Our approach relies on two complementary components to address the functional misalignment of MLLMs and the limitations of negative mining.

\subsection{Preliminaries: Multimodal Embedding with MLLMs}

The primary goal of a universal multimodal embedding model is to establish a unified vector space capable of handling diverse downstream tasks across heterogeneous modalities, such as text, images, or interleaved inputs. To achieve this, our framework is built upon a pre-trained Multimodal Large Language Model (MLLM), denoted as $f_\theta$. A typical MLLM synergistically integrates three core components: (1) a vision encoder that processes visual information, (2) a projection layer that aligns image features into the linguistic space, and (3) an LLM backbone that processes the unified sequence of visual and textual tokens to handle task-specific instructions.

Formally, let $\mathcal{B} = \{(q_i, c_i, \mathcal{I}_i)\}_{i=1}^{N}$ denote a batch of $N$ training triplets, where $q_i$ represents a query, $c_i$ is the corresponding positive candidate, and $\mathcal{I}_i$ is the task instruction. Given an input $x$ (either $q$ or $c$) and an instruction $\mathcal{I}$, the MLLM processes them as a continuous stream of tokens. Following existing approaches~\cite{jiang2025vlmvec,jiang2024e5,li2025making}, we extract the representation of the last token from the last hidden layer as the unified embedding vector (\textit{i.e.}, last token pooling): $\mathbf{e} = f_\theta(x, \mathcal{I})$. Here, we explicitly define candidate $c$ as a universal retrieval-side object across heterogeneous tasks; task-specific outputs such as textual answers in VQA, class labels in Classification, target documents in Retrieval, or visual regions in Grounding are all uniformly treated as candidates in the shared embedding space.

The similarity between a query $q$ and a candidate $c$ is measured using cosine similarity:
\begin{equation}
  s(q, c) = \frac{\mathbf{e}_q^\top \mathbf{e}_c}{\|\mathbf{e}_q\| \|\mathbf{e}_c\|}.
\end{equation}

To optimize this embedding space, we adopt the standard contrastive learning objective, which pulls positive pairs together while pushing apart all other samples in the batch. The loss for the $i$-th query is formulated as:
\begin{equation}
\label{eq:infonce}
  \mathcal{L}_{\text{InfoNCE}} = - \log \frac{\exp(s(q_i, c_i) / \tau)}{\exp(s(q_i, c_i) / \tau) + \sum_{j \neq i} \exp(s(q_i, c_j) / \tau)},
\end{equation}

\noindent where $\tau$ is a temperature parameter.

\subsection{Latent Conditioning via Hierarchical Prompting}
\label{sec:prompt}

\begin{table}[t]
\caption{\textbf{Impact of prompting strategies on alignment and modality gap.} We compare standard user prompts (REP) against generic (SYS-Non) and task-specific (SYS-T) system instructions. The modality gap scores are evaluated on the MSCOCO dataset using the Qwen2-VL backbone.}
\label{tab:prompt_analysis}
\centering
\resizebox{.97\columnwidth}{!}{%
\begin{tabular}{lcccc}
\toprule
\multirow{2}{*}{Prompt} & \multicolumn{2}{c}{Alignment $\uparrow$} & \multicolumn{2}{c}{Modality Gap $\downarrow$} \\
\cmidrule(lr){2-3} \cmidrule(lr){4-5}
 & 2B & 7B & 2B & 7B \\ 
\midrule
- & 0.731 & 0.836 & 0.556 & 0.362 \\
SYS-Non & 0.711 & 0.734 & 0.604 & 0.567 \\ \midrule
REP & \cellcolor{blue!18}0.893 & \cellcolor{blue!7}0.893 & \cellcolor{red!33}0.184 & \cellcolor{red!50}0.270 \\
SYS-Non + REP & \cellcolor{blue!5}0.881 & \cellcolor{blue!5}0.890 & \cellcolor{red!50}0.211 & \cellcolor{red!32}0.202 \\
SYS-T & \cellcolor{blue!38}0.911 & \cellcolor{blue!50}0.954 & \cellcolor{red!15}0.154 & \cellcolor{red!5}0.095 \\
SYS-T + REP-IMG & \cellcolor{blue!45}0.917 & \cellcolor{blue!49}0.953 & \cellcolor{red!11}0.147 & \cellcolor{red!8}0.110 \\
SYS-T + REP-TXT & \cellcolor{blue!34}0.907 & \cellcolor{blue!49}0.953 & \cellcolor{red!31}0.180 & \cellcolor{red!6}0.099 \\
SYS-T + REP & \cellcolor{blue!50}0.921 & \cellcolor{blue!48}0.952 & \cellcolor{red!5}0.137 & \cellcolor{red!8}0.108 \\
\bottomrule
\end{tabular}%
}
\vspace{-6pt}
\end{table}

Unlike contrastive models such as CLIP, MLLMs are not natively trained to represent multimodal inputs as unified embeddings. This absence of explicit alignment training often results in a significant modality gap, where visual and textual representations remain distinct in the latent space. Recent studies~\cite{ouali2025vladva,jiang2024e5} demonstrate that explicitly instructing the model with ``in one word'' can effectively bridge this gap by unifying multimodal inputs into a shared linguistic space. Although these works highlight the efficacy of prompting, the structural impact of system-level conditioning remains underexplored. We investigate this through a controlled analysis of prompting strategies, inspired by the modality gap analysis in contrastive learning~\cite{liang2022mind}.

\subsubsection{The Role of Task-Specific Constraints}

We first analyze whether the mere presence of a system instruction improves alignment. As shown in Table~\ref{tab:prompt_analysis}, a generic system prompt (SYS-Non, \textit{e.g.}, ``You are a helpful assistant.") fails to reduce the modality gap compared to the baseline. This indicates that the alignment is not driven by the system prompt structure itself, but by the specific task constraint (\textit{i.e.}, ``in one word") contained within it.

\subsubsection{System-level vs. User-level Conditioning}

When the task constraint is applied, its placement becomes crucial. Comparing user-level instructions (REP) against system-level instructions (SYS-T), we observe that placing the constraint in the system prompt yields superior alignment. By defining the task at the system level, the model is conditioned with a strong task prior before processing the variable user input. This global conditioning proves more effective than providing the instruction solely as a local user directive (REP), which acts locally within the input context.

\subsubsection{Asymmetric Reinforcement}

Building on the necessity of system-level constraints, we investigate their optimal application across queries and candidates. Table~\ref{tab:ablation_prompt_strategies} demonstrates that applying the system prompt symmetrically (System-QD) is critical, dramatically improving average performance from 14.2 to 41.7. This confirms that a shared global anchor is essential for initial alignment.

However, for user-level reinforcement, an asymmetric approach is optimal. Naively adding user instructions to both sides (+ QD-Rein.) degrades overall performance. This observation aligns with recent findings in LLM-based text embeddings~\cite{li2025making}, which demonstrate that applying prompts to the candidate (passage) side severely harms performance across non-retrieval tasks. We attribute this to a functional divergence: candidates require pure information compression, making extra local instructions act as disruptive noise. In contrast, queries possess complex intents (\textit{e.g.}, task-specific questions) that benefit from explicit reinforcement.

Consequently, applying user-level reinforcement solely to the query (+ Q-Rein.) achieves the highest overall performance. This asymmetric strategy significantly enhances intent-heavy tasks like Classification and VQA while preserving Retrieval accuracy. Therefore, we formalize our optimal conditioning: $\mathcal{I}_{doc}=[\mathcal{I}_{sys};x]$ for pure compression, and $\mathcal{I}_{query}=[\mathcal{I}_{sys};\mathcal{I}_{user};x]$ for fine-grained intent capture. These exact hierarchical templates are explicitly detailed in Appendix B. This structure provides a stable initialization, enabling SaHa to function effectively from the early stages. We extensively demonstrate the universal applicability of this latent conditioning across diverse architectures and benchmarks in Section \ref{sec:exp}.

\begin{table}
\centering
\caption{\textbf{Ablation on prompt strategies using the MMEB benchmark.} This study validates the effectiveness of different prompt combinations in a real-world setting.}
\label{tab:ablation_prompt_strategies}
\resizebox{.9\columnwidth}{!}{%
\begin{tabular}{lccccc}
\toprule
Strategy & Cla. & VQA & Ret. & Gro. & Avg. \\ \midrule
No-Prompt   & 21.0 & 8.7  & 9.1 & 26.1 & 14.2 \\ \midrule
System-Q  & 21.6 & 7.5  & 13.0 & 12.1 & 13.8 \\ 
System-D  & 35.8 & 25.4 & 20.6 & 25.3 & 26.7 \\ \midrule
System-QD & 46.7 & 31.8 & \textbf{44.1} & 46.7 & 41.7 \\
\text{  } + QD-Rein. & 45.8 & 30.3 & 37.1 & 46.7 & 39.1 \\
\text{  } + D-Rein. & 46.2 & \textbf{33.3} & 41.9 & 38.6 & 40.9 \\
\text{  } + Q-Rein. & \textbf{48.6} & \textbf{33.3} & \textbf{44.1} & \textbf{52.6} & \textbf{43.3} \\
\bottomrule
\end{tabular}
}
\vspace{-6pt}
\end{table}

\begin{figure*}
\centering
\includegraphics[width = .99\textwidth]{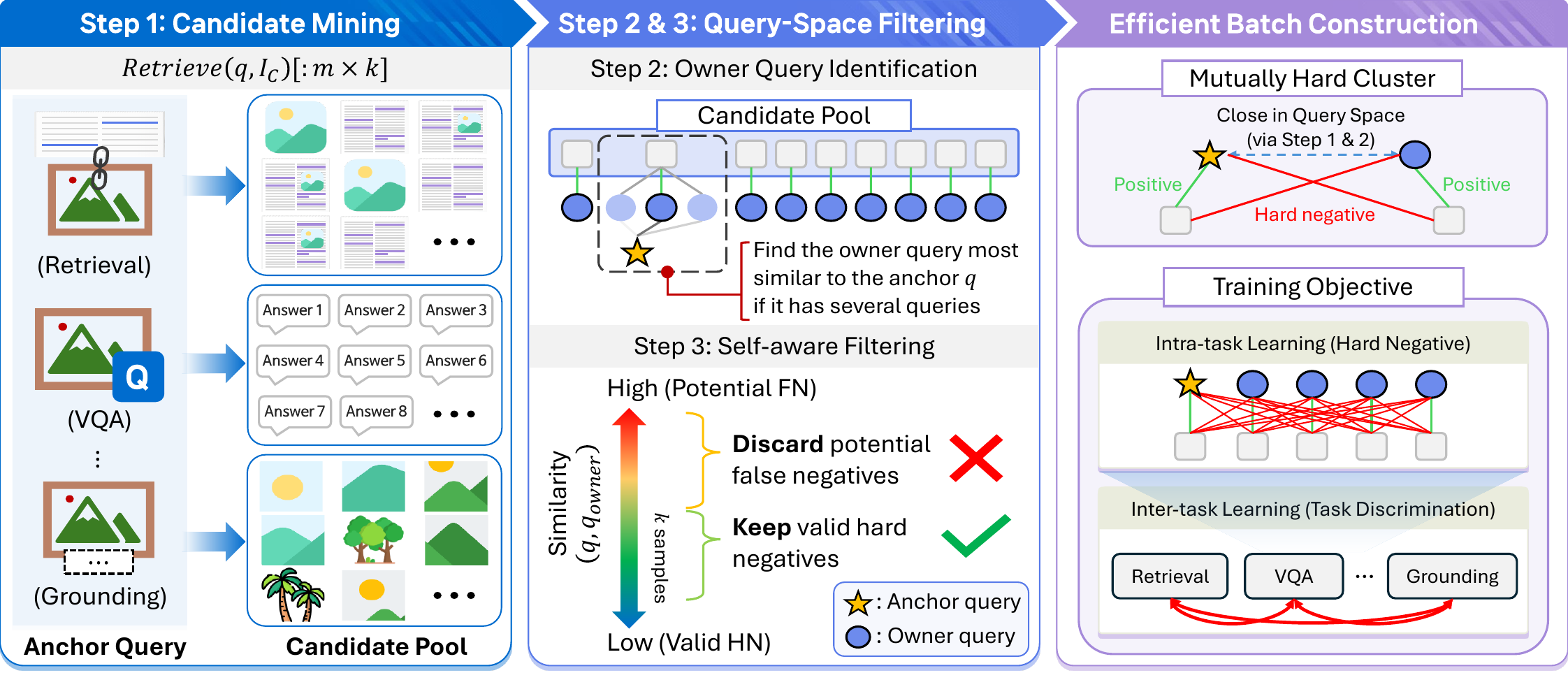}
\captionof{figure}{
    \textbf{Overview of the Self-aware Hard Negative Sampling (SaHa) strategy.} We first mine a broad candidate pool ($m \times k$) for a given anchor query. To prevent false negatives, candidates are mapped to their owner queries. By measuring similarity against the anchor query, we discard high-similarity pairs as potential false negatives and select the $k$ least similar ones as valid hard negatives. The selected samples form mutually hard clusters, optimizing both fine-grained intra-task discrimination and global inter-task separation within a single efficient batch.}
\vspace{-6pt}
\label{fig:saha_method}
\end{figure*}

\subsection{Self-aware Hard Negative Sampling}

The hierarchical prompt establishes a well-aligned foundational space for diverse embedding tasks, but optimizing the model for fine-grained discrimination requires hard negative mining. Traditional hard negative sampling (HNS) retrieves candidates that are semantically close to an anchor query $q$. However, this purely candidate-centric approach inevitably retrieves unlabeled positive candidates, leading to severe false negative contamination. Previous methods attempt to mitigate this by relying on external teacher models or fragile similarity thresholds~\cite{gu2025breaking}, which are computationally expensive and highly sensitive to dataset variations.

To overcome these limitations, we introduce Self-aware Hard Negative Sampling (SaHa), a novel mining strategy that fundamentally shifts the filtering mechanism from candidate-space to query-space. By leveraging the model's own structurally conditioned embeddings, SaHa identifies valid hard negatives and constructs highly efficient, mutually challenging training clusters without any external dependencies. 

\subsubsection{Query-Space Filtering for False Negative Prevention}

The core novelty of SaHa lies in our intuition regarding latent semantic structures: \textit{semantically similar queries are highly likely to share similar positive candidates}. Therefore, if a retrieved hard candidate belongs to a query that is semantically identical to the anchor, that candidate is almost certainly a false negative. Based on this intuition, SaHa operates in three distinct steps to select $k$ valid hard negatives for a given anchor query $q_i$, as illustrated in Fig.~\ref{fig:saha_method}:

\textbf{Step 1: Candidate Mining.} We retrieve an initial challenging candidate pool $D_{\mathrm{cand}}(q_i)$ from the global candidate space $\mathcal{C}$ using cosine similarity:
\begin{equation}
  D_{\mathrm{cand}}(q_i) = \mathrm{Top}(m \times k, \{c \in \mathcal{C} \mid s(q_i, c)\})
\end{equation}
where $m$ is the pool multiplier and $k$ is the target number of hard negatives.

\textbf{Step 2: Owner Query Identification.} A single candidate might act as a positive target for multiple queries across the dataset $\mathcal{B}_{\mathrm{full}}$. We first define the owner-query set $\mathrm{Owner}(c)$ as all training queries associated with candidate $c$:
\begin{equation}
  \mathrm{Owner}(c) = \{q \mid (q, c) \in \mathcal{B}_{\mathrm{full}}\}
\end{equation}
To rigorously evaluate the false-negative risk, we identify the selected owner query $q_{\mathrm{owner}}(c, q_i)$ that is most semantically similar to the anchor $q_i$:
\begin{equation}
  q_{\mathrm{owner}}(c, q_i) = \mathop{\arg\max}_{q \in \mathrm{Owner}(c)} s(q_i, q)
\end{equation}

\textbf{Step 3: Self-aware Filtering.} To strictly filter out semantic false negatives, we compute the similarity $s(q_i, q_{\mathrm{owner}}(c, q_i))$ between the anchor and the selected owner queries. The final set of $k$ true hard negatives $\mathcal{N}(q_i)$ is constructed by selecting a subset $\mathcal{C}_{neg}$ whose owner queries exhibit the lowest semantic overlap with the anchor:
\begin{equation}
  \mathcal{N}(q_i) = \mathop{\arg\min}_{\mathcal{C}_{neg} \subset D_{\mathrm{cand}}(q_i), |\mathcal{C}_{neg}|=k} \sum_{c \in \mathcal{C}_{neg}} s(q_i, q_{\mathrm{owner}}(c, q_i))
\end{equation}
This formalization clarifies that \textit{hardness} comes from retrieving candidates similar to the anchor query (Step 1), whereas \textit{safety} comes from filtering candidates whose owner queries are semantically close to the anchor (Step 3). Notably, for tasks with explicit categorical ground truths (\textit{e.g.}, Classification), we bypass this similarity measurement and directly discard candidates whose owner queries share the same class label as the anchor, ensuring absolute precision in false negative removal.

By selecting candidates owned by distinct queries, we guarantee that the sampled candidates are hard (due to Step 1) yet valid true negatives (due to Step 3). This offline pre-processing strategy dynamically adapts to any dataset, avoiding the need for brittle hyperparameter tuning. Note that SaHa serves exclusively as a one-time offline preprocessing strategy for training data curation. Consequently, during inference, the fine-tuned model generates embeddings via a standard forward pass without incurring any additional clustering overhead or latency. Since SaHa is a one-time, offline preprocessing step that operates only on the training set, it adds no overhead to training or inference. In practice, on 6 NVIDIA RTX A6000 GPUs, the full MMEB-train pipeline takes roughly 10 hours to extract candidate embeddings and an additional $\sim$1.3 hours for similarity-based hard-negative mining. With peak memory bounded by a single $|C|\times d$ embedding matrix (where $d$ is the embedding dimension), the dominant per-task computation scales as $O(|Q|\cdot|C|)$ where $|Q|$ and $|C|$ are the total number of queries and candidates, respectively. However, this process is highly optimized depending on the task; for instance, classification tasks finish in minutes because their candidate pool collapses to the exact label set. Ultimately, our procedure efficiently scales sub-linearly to larger candidate pools via ANN indexing, demonstrating high scalability to large-scale datasets without acting as a computational bottleneck. The detailed pseudocode of SaHa, including our two-phase allocation strategy to ensure all samples are assigned without excessive redundancy, is provided in Appendix.

\subsubsection{Batch Efficiency via Mutually Hard Clusters}

Beyond preventing false negatives, SaHa introduces a structural innovation that maximizes training efficiency. Traditional HNS creates a severe computational bottleneck by appending $k$ isolated negative candidates for every query, requiring redundant forward passes.

SaHa, in contrast, constructs tailored mutually hard clusters. For a given anchor $q_i$, we group its positive pair $(q_i, c_i^+)$ with the $k$ selected hard negative pairs $\{(q_j, c_j^+)\}_{j=1}^k$. Within this cluster of size $k+1$, every candidate serves a dual role: it acts as a labeled positive for its owner query, while simultaneously serving as a rigorous hard negative for the other queries in the cluster. This dense relational structure ensures that every forward pass contributes directly to fine-grained discrimination, improving batch efficiency.

\subsubsection{Training Objective}

During fine-tuning, we optimize the model using the standard InfoNCE objective defined in Eq.~\ref{eq:infonce}. However, rather than using a randomly sampled batch, we compute it strictly within the SaHa-constructed clusters. For a given anchor $q_i$, the negative pool is explicitly restricted to the $k$ positive candidates $\{c_j^+\}_{j \neq i}$ belonging to the other queries in the same cluster. By processing $B$ such tailored clusters in parallel, our effective training batch size becomes $B \times (k+1)$. This ensures that every contrastive comparison directly contributes to fine-grained, intra-task discrimination without incurring additional computational overhead.

\section{Experiments}
\label{sec:exp}

\subsection{Experimental Setup}
\label{subsec:setup}

\begin{table*}
\centering
\caption{
    \textbf{Comparison with state-of-the-art methods on the MMEB benchmark.} 
    Cla., VQA, Ret., Gro., IND, and OOD stand for Classification, VQA, Retrieval, Grounding, In-Domain, and Out-of-Domain, respectively. The highest score in each column is highlighted in bold. The second-best result is underlined.
}
\label{tab:main_results}
\resizebox{.9\textwidth}{!}{%
\begin{tabular}{lccccccccc}
\toprule
\multirow{2}{*}{Method} & \multirow{2}{*}{Model} & \multirow{2}{*}{\#Params} & \multicolumn{4}{c}{Per Meta-Task Score } & \multicolumn{3}{c}{Average Score} \\
\cmidrule(lr){4-7} \cmidrule(lr){8-10}
 &  &  & Classification & VQA & Retrieval & Grounding & IND & OOD & Overall \\
\midrule
\#Datasets $\rightarrow$ & & & 10 & 10 & 12 & 4 & 20 & 16 & 36 \\
\midrule
\rowcolor{blue!3}\multicolumn{10}{c}{\textit{Partially Supervised Finetuning Setting (Finetuning on M-BEIR)}} \\
\midrule
GME~\cite{zhang2025bridging} & Qwen2-VL & 2.2B & 56.9 & 41.2 & 67.8 & 53.4 & - & - & 55.8 \\
MM-Embed~\cite{mmembed} & LLaVA-1.6 & 7.8B & 48.1 & 32.3 & 63.8 & 57.8 & - & - & 50.0 \\
\midrule
\rowcolor{blue!3}\multicolumn{10}{c}{\textit{Supervised Finetuning Setting (Finetuning on MMEB, $<$ 5B params)}} \\
\midrule
OpenCLIP~\cite{cherti2023reproducible} & CLIP & 0.4B & 56.0 & 21.9 & 55.4 & 64.1 & 50.5 & 43.1 & 47.2 \\  
E5-V~\cite{jiang2024e5} & Phi3.5-V & 4.2B & 39.1 & 9.6 & 38.0 & 57.6 & 33.1 & 31.9 & 32.6 \\
VLM2Vec~\cite{jiang2025vlmvec} & Qwen2-VL & 2.2B & 59.0 & 49.4 & 65.4 & 73.4 & 66.0 & 52.6 & 59.3 \\
VLM2Vec~\cite{jiang2025vlmvec} & Phi-3.5-V & 4.2B & 54.8 & 54.9 & 62.3 & 79.5 & 66.5 & 52.0 & 60.1 \\
UniME~\cite{gu2025breaking} & Phi-3.5-V & 4.2B & 54.8 & 55.9 & 64.5 & 81.8 & 68.2 & 52.7 & 64.2 \\
CAFe~\cite{yu2025cafe} & LLaVA-OV~\cite{li2025llavaonevision} & 0.9B & 59.1 & 49.1 & 61.0 & 83.0 & 64.3 & 53.7 & 59.6 \\
UNITE~\cite{kong2025modality} & Qwen2-VL & 2.2B & 63.2 & 55.9 & 65.4 & 75.6 & 65.8 & 60.1 & 63.3 \\
LLaVE~\cite{lan2025llave} & Aquila-VL~\cite{gu2024infinity} & 2.0B & 62.1 & \textbf{60.2} & 65.2 & \underline{84.9} & 69.4 & 59.8 & 65.2 \\
\rowcolor{gray!8}
\textbf{Ours} & LFM2.5-VL~\cite{amini2025lfm2} & 1.6B & \underline{64.6} & \underline{59.6} & \underline{67.5} & \textbf{87.0} & \textbf{70.6} & \underline{61.7} & \underline{66.6} \\
\rowcolor{gray!8}
\textbf{Ours} & Qwen2-VL & 2.2B & \textbf{65.8} & 59.4 & \textbf{70.2} & 82.6 & \underline{70.5} & \textbf{63.4} & \textbf{67.4} \\
\midrule
\rowcolor{blue!3}\multicolumn{10}{c}{\textit{Supervised Finetuning Setting (Finetuning on MMEB, $>$ 5B params)}} \\
\midrule
VLM2Vec~\cite{jiang2025vlmvec} & LLaVA-1.6 & 7.6B & 61.2 & 49.9 & 67.4 & 86.1 & 67.5 & 57.1 & 62.9 \\
VLM2Vec~\cite{jiang2025vlmvec} & Qwen2-VL & 8.3B & 62.6 & 57.8 & 69.9 & 81.7 & 72.2 & 57.8 & 65.8 \\
MMRet~\cite{zhou2024megapairs} & LLaVA-1.6 & 7.6B & 56.0 & 57.4 & 69.9 & 83.6 & 68.0 & 59.1 & 64.1 \\
UniME~\cite{gu2025breaking} & LLaVA-1.6 & 7.6B & 60.6 & 52.9 & 67.9 & 85.1 & 68.4 & 57.9 & 66.6 \\
CAFe~\cite{yu2025cafe} & LLaVA-OV~\cite{li2025llavaonevision} & 8.0B & 65.2 & 65.6 & 70.0 & 91.2 & 75.8 & 62.4 & 69.8 \\
UNITE~\cite{kong2025modality} & Qwen2-VL & 8.3B & 68.3 & 65.1 & 71.6 & 84.8 & 73.6 & 66.3 & 70.3 \\
LLaVE~\cite{lan2025llave} & Aquila-VL~\cite{gu2024infinity} & 8.0B & 65.7 & 65.4 & 70.9 & \textbf{91.9} & 75.0 & 64.4 & 70.3 \\
\rowcolor{gray!8}
\textbf{Ours} & Qwen2-VL & 8.3B & \textbf{69.1} & \textbf{67.3} & \textbf{74.1} & 88.1 & \textbf{76.4} & \textbf{67.4} & \textbf{72.4} \\
\bottomrule
\end{tabular}%
}
\vspace{-6pt}
\end{table*}

\subsubsection{Implementation Details}
Our experiments are primarily built upon the Qwen2-VL~\cite{wang2024qwen2} architecture. Unless otherwise specified, ablations and analyses are conducted on the 2.2B model (Qwen2-VL-2B) for efficiency, though our framework is strictly model-agnostic. For fine-tuning, we use the AdamW optimizer~\cite{loshchilov2018decoupled} with a learning rate of $5 \times 10^{-5}$ and a linear schedule with a 100-step warm-up. The temperature $\tau$ and maximum sequence length are set to 0.02 and 1024 tokens, respectively. The model is trained for 1.5 epochs using a LoRA adapter (rank 8). While input images are resized to a fixed $448 \times 448$ resolution during training, we allow for dynamic resolutions at inference. To increase the effective batch size beyond hardware limitations, we employ GradCache~\cite{gao-etal-2021-scaling}, a gradient caching method. All training experiments were conducted on 8 RTX A6000 GPUs.

\subsubsection{Datasets}
We evaluate our model on the Massive Multimodal Embedding Benchmark (MMEB)~\cite{jiang2025vlmvec}, comprising 36 datasets across four meta-tasks: Classification (Cla.), Visual Question Answering (VQA), Retrieval (Ret.), and Visual Grounding (Gro.). For fine-tuning, we use the combined training sets from 20 in-domain datasets ($\sim$829K pairs). The remaining 16 out-of-domain datasets are used exclusively for zero-shot evaluation. We report Hit@1 for the MMEB benchmark. Furthermore, to explicitly assess cross-modality generalization, we utilize six representative video datasets following the MMEBv2~\cite{meng2026vlmvecv} evaluation protocol. All tasks are evaluated in an embedding-based retrieval manner. Detailed statistics and descriptions of these datasets are provided in Appendix.

\subsection{Main Results}

\subsubsection{Universal Embedding Performance}
Table~\ref{tab:main_results} compares our framework against state-of-the-art methods on the MMEB benchmark. Our fine-tuned 2.2B model achieves a strong overall score of 67.4, setting state-of-the-art performance in its parameter class for both Classification and Retrieval tasks. This effectiveness scales seamlessly: our 8.3B model pushes the overall score to 72.4, demonstrating dominant performance across Classification, VQA, and Retrieval. Notably, our framework excels in the Retrieval meta-task, a significant challenge for generative MLLMs, validating SaHa's capability to mine highly discriminative negatives. Furthermore, to verify architectural universality, we extended our framework to the Liquid Foundation Model (LFM2.5-VL~\cite{amini2025lfm2}), an emerging efficient architecture. Applying our hierarchical prompting and SaHa to LFM2.5-VL yields consistent SOTA-level gains, confirming that our method's efficacy stems from fundamental latent space alignment rather than the specific inductive biases of standard Transformers.

\begin{table*}[t!]
\centering
\caption{\textbf{Zero-shot results on SugarCrepe and SugarCrepe++ compositionality benchmark.}}
\label{tab:my_results}
\resizebox{.9\textwidth}{!}{%
\begin{tabular}{@{}l c c c c c c c c c c c c@{}}
\toprule
\multirow{3}{*}{Method} & \multirow{3}{*}{Params} & \textbf{SugarCrepe} & \multicolumn{10}{c}{\textbf{SugarCrepe++}} \\
& & \multirow{2}{*}{Avg.} & \multicolumn{2}{c}{Swap Object} & \multicolumn{2}{c}{Swap Attribute} & \multicolumn{2}{c}{Replace Object} & \multicolumn{2}{c}{Replace Attribute} & \multicolumn{2}{c}{Replace Relation} \\
\cmidrule(lr){4-5} \cmidrule(lr){6-7} \cmidrule(lr){8-9} \cmidrule(lr){10-11} \cmidrule(lr){12-13}
& & & ITT & TOT & ITT & TOT & ITT & TOT & ITT & TOT & ITT & TOT \\
\midrule
Human & - & 99.0 & 100.0 & 96.7 & 96.7 & 93.3 & 100.0 & 97.0 & 100.0 & 98.3 & 100.0 & 96.7 \\ \midrule
\rowcolor{blue!3}\multicolumn{13}{c}{\textit{Training-Free Prompting}} \\ \midrule
VladVA (Qwen2-VL-2B) & 2.21B & 70.3 & 32.7 & 27.8 & 30.5 & 25.3 & 73.6 & 65.9 & 46.8 & 43.0 & \underline{57.6} & \textbf{58.3} \\
VladVA (LLaVA-1.5-7B) & 7.06B & 70.7 & 23.8 & 30.7 & 28.0 & 29.5 & 58.1 & 63.0 & 46.8 & 58.1 & 52.3 & 63.4 \\
\rowcolor{gray!8} Ours (Qwen2-VL-2B) & 2.21B & \underline{76.6} & \underline{41.2} & \underline{31.0} & \underline{36.8} & \underline{30.2} & \textbf{84.0} & \underline{81.4} & \textbf{61.5} & \textbf{63.8} & 54.5 & 54.8 \\
\rowcolor{gray!8} Ours (Qwen2-VL-7B)    & 8.21B & \textbf{77.6} & \textbf{44.1} & \textbf{35.9} & \textbf{40.8} & \textbf{38.7} & 83.4 & \textbf{85.7} & \underline{59.5} & \underline{60.0} & \textbf{58.8} & \underline{58.1} \\
\midrule
\rowcolor{blue!3}\multicolumn{13}{c}{\textit{Contrastive Models ($<$ 3B params)}} \\ \midrule
OpenCLIP (ViT-G/14)   & 1.37B & 80.1 & 40.7 & 27.4 & 54.2 & 49.6 & 93.1 & 89.4 & 72.5 & 73.1 & 57.6 & 51.4 \\
OpenCLIP (ViT-BigG/14) & 2.54B & 82.0 & \underline{48.8} & 28.2 & 57.7 & \underline{52.4} & \underline{94.2} & 90.5 & \underline{76.4} & 72.6 & 59.4 & 53.6 \\
VladVA (Qwen2-VL-2B) & 2.21B & \textbf{88.2} & \textbf{50.8} & \underline{33.5} & \textbf{60.4} & 48.2 & 93.7 & \underline{93.8} & 74.8 & \underline{77.5} & \underline{63.6} & \underline{57.4} \\
\rowcolor{gray!8} Ours (Qwen2-VL-2B) & 2.21B & \underline{87.0} & 46.1 & \textbf{38.4} & \underline{59.5} & \textbf{55.4} & \textbf{95.6} & \textbf{96.5} & \textbf{81.9} & \textbf{85.5} & \textbf{72.0} & \textbf{64.4} \\
\midrule
\rowcolor{blue!3}\multicolumn{13}{c}{\textit{Contrastive Models ($>$ 3B params)}} \\ \midrule
E5-V (LLaVA-1.5-7B) & 7.06B & 81.8 & 39.5 & 42.3 & 40.7 & 48.5 & 89.7 & 94.6 & 71.7 & \underline{86.4} & 72.0 & \underline{81.5} \\
E5-V (LLaVA-Next-8B) & 8.36B & 84.2 & 50.8 & \underline{48.4} & 49.7 & 56.9 & 93.1 & \textbf{97.6} & 76.1 & \textbf{87.1} & \textbf{74.7} & \textbf{84.4} \\
VLM2Vec (LLaVA-1.6-7B) & 7.30B & 83.8 & 40.7 & 39.9 & 48.1 & 50.0 & 94.6 & \underline{96.9} & 77.0 & 85.6 & 67.9 & 70.7 \\
VladVA (LLaVA-1.5-7B) & 7.06B & \textbf{90.0} & \textbf{56.1} & 36.7 & \underline{63.0} & \underline{62.5} & \textbf{95.0} & 93.0 & \underline{78.2} & 82.3 & 71.1 & 66.3 \\
\rowcolor{gray!8} Ours (Qwen2-VL-7B) & 8.21B & \underline{87.5} & \underline{51.0} & \textbf{51.0} & \textbf{73.1} & \textbf{77.8} & 94.2 & 93.4 & \textbf{82.2} & 83.1 & \underline{73.1} & 71.8 \\
\bottomrule
\end{tabular}%
}
\vspace{-6pt}
\label{tab:sugar_plus}
\end{table*}

\subsubsection{Fine-grained Compositionality}
To evaluate fine-grained visual-linguistic alignment, we assess the model on the SugarCrepe~\cite{hsieh2023sugarcrepe} and SugarCrepe++~\cite{dumpala2024sugarcrepe} benchmarks, which are specifically designed to test a model's ability to understand subtle semantic changes in image captions, such as swapping objects or attributes. As shown in Table~\ref{tab:sugar_plus}, in the training-free setting, our hierarchical prompting alone significantly surpasses standard user-level instruction baselines (\textit{e.g.}, VladVA) across all sizes. More importantly, after contrastive fine-tuning, our approach achieves highly competitive or superior performance—particularly in the challenging `Swap Attribute' and `Replace Attribute' categories—despite utilizing only a fraction ($\sim$10\%) of the domain-specific training data (0.8M multi-task pairs) compared to large-scale baselines trained on over 8M image-caption pairs. This highlights the exceptional data efficiency and compositional understanding unlocked by our framework.

\begin{table}[t]
\centering
\caption{\textbf{Zero-shot cross-modality generalization.} We report Hit@1 performance on all video tasks. All methods were built from the Qwen2-VL-2B model.}
\label{tab:video_generalization}
\begin{tabular}{lccc}
\toprule
\textbf{Video Task} & GME~\cite{zhang2025bridging} & VLM2Vec~\cite{jiang2025vlmvec} & \textbf{Ours} \\
\midrule
ActivityNetQA & 58.0 & 49.6 & \textbf{71.1} \\
Breakfast & 13.6 & 13.4 & \textbf{20.1} \\
EgoSchema & \textbf{40.8} & 25.4 & 25.0 \\ 
Kinetics-700 & 35.2 & 31.4 & \textbf{42.2} \\
UCF101 & 52.4 & 57.5 & \textbf{68.4} \\
Video-MME & \textbf{34.3} & 26.9 & 21.6 \\ 
\midrule
Average & 39.1 & 34.0 & \textbf{41.4} \\
\bottomrule
\end{tabular}%
\end{table}

\subsubsection{Cross-Modality Generalization to Video}
To directly address the universality of our learned embedding space beyond static images, we perform a zero-shot cross-evaluation on the unseen video modality. Following the MMEBv2 protocol, we evaluate our model, fine-tuned exclusively on spatial image-text pairs, on six diverse video datasets encompassing action recognition and complex temporal reasoning.

\begin{figure}
\centering
\includegraphics[width=1.\columnwidth]{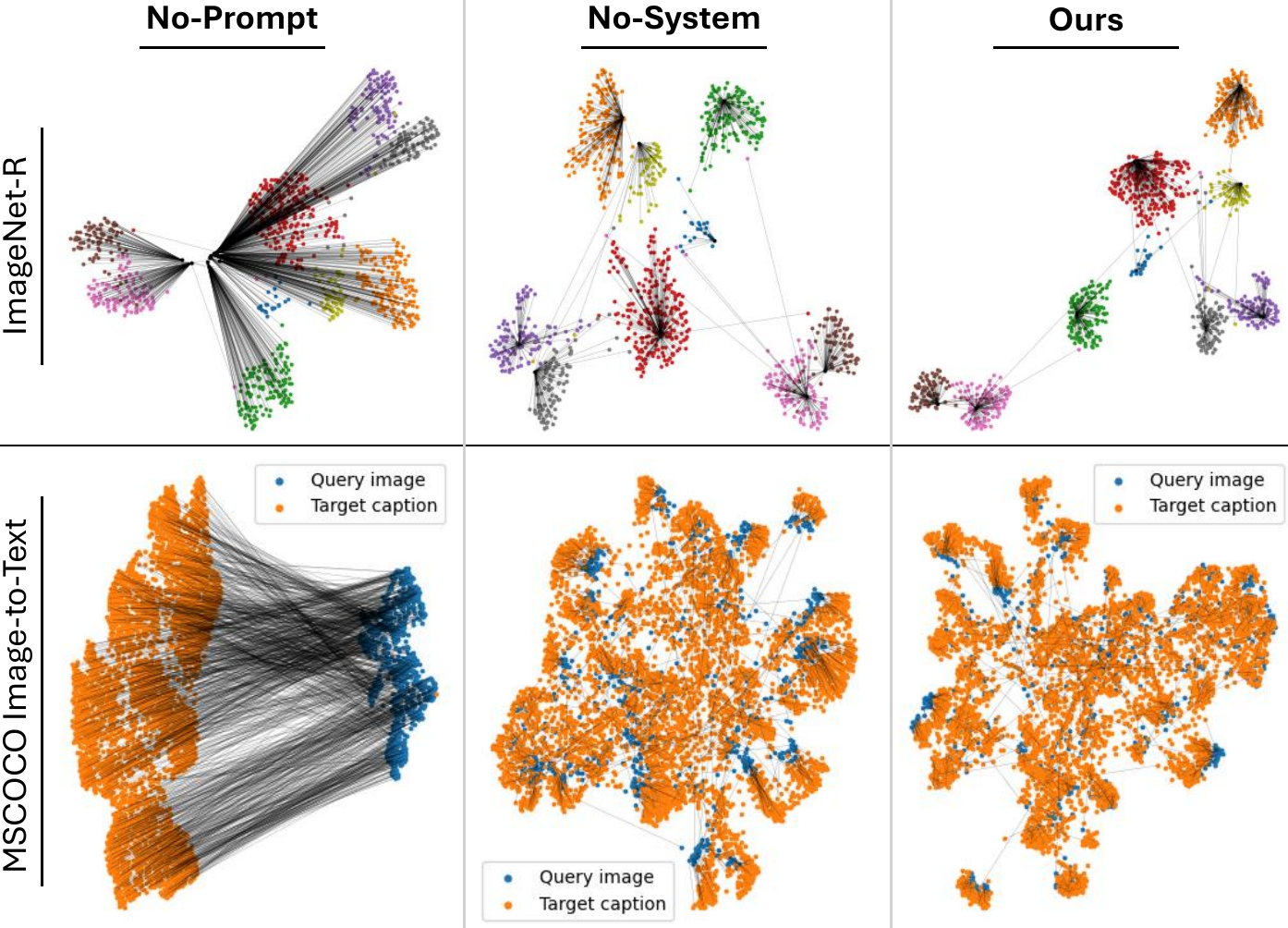}
\caption{\textbf{UMAP visualization of the zero-shot embedding space.}}
\vspace{-6pt}
\label{fig:umap_visualization}
\end{figure}

As detailed in Table \ref{tab:video_generalization}, we compare our method against recent state-of-the-art multimodal embedding models, GME~\cite{zhang2025bridging} and VLM2Vec~\cite{jiang2025vlmvec}, with all methods utilizing the same Qwen2-VL-2B backbone. Our framework successfully extracts meaningful temporal representations without any video-specific training or temporal pooling heuristics, achieving the highest average score of 41.4\%. We observe particularly strong gains on action-centric and temporally extended QA tasks such as ActivityNetQA (71.1\%) and UCF101 (68.4\%). This compelling result substantiates that our unified approach establishes a fundamentally robust and extensible multimodal representation space capable of seamlessly bridging multiple modalities.

\subsection{Deep Analysis of Our Framework}

\subsubsection{Impact of Latent Conditioning}
\label{subsec:prompting_results}

\begin{figure}
\centering
\includegraphics[width = .98\columnwidth]{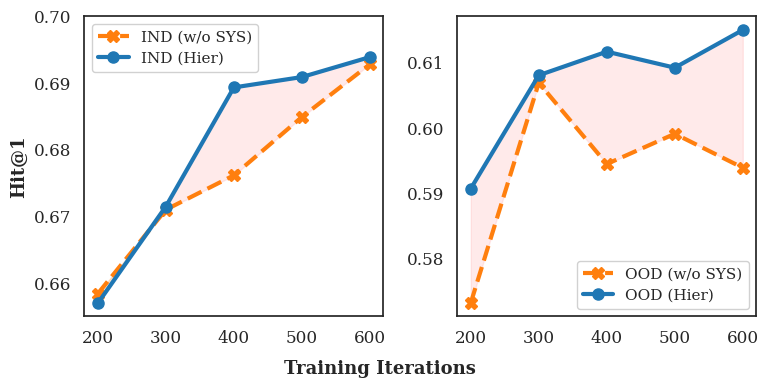}
\captionof{figure}{
    \textbf{Training dynamics and robustness analysis.} We compare the Precision@1 scores of the standard user-prompt baseline (w/o SYS) and our hierarchical approach (Hier) across training iterations.}
\vspace{-6pt}
\label{fig:training_dynamics}
\end{figure}

\begin{table}[]
\caption{\textbf{Impact of latent conditioning across diverse architectures.} Our training-free method, which solely relies on our hierarchical prompt, is compared against various contrastively trained vision-language models (VLMs) and MLLMs.}
\label{tab:zeroshot_mmeb}
\resizebox{\columnwidth}{!}{%
\begin{tabular}{lcccccc}
\toprule
Method & \#Param & Cla. & VQA & Ret. & Gro. & Avg. \\  \midrule
\multicolumn{7}{c}{Contrastive VLMs} \\ \midrule
CLIP~\cite{radford2021learning} & 0.4B & 42.8 & 9.1 & 53.0 & 51.8 & 37.8 \\
OpenCLIP~\cite{chcherti2023reproducible} & 0.4B & 47.8 & 10.9 & 52.3 & 53.3 & 39.7 \\
SigLIP~\cite{zhai2023sigmoid} & 0.9B & 40.3 & 8.4 & 31.6 & 59.5 & 34.8 \\
MagicLens~\cite{magiclens} & 0.4B & 38.8 & 8.3 & 35.4 & 26.0 & 27.1 \\ \midrule
\multicolumn{7}{c}{Contrastive MLLMs} \\ \midrule
E5-V~\cite{jiang2024e5} (LLaVA-NeXT~\cite{li2025llavanextinterleave}) & 8.4B & 21.8 & 4.9 & 11.5 & 19.0 & 13.3 \\
UniME~\cite{gu2025breaking} (Phi3.5-V~\cite{abdin2024phi}) & 4.2B & 42.5 & 18.3 & 40.5 & 59.9 & 40.3 \\ 
MMRet~\cite{zhou2024megapairs} (LLaVA-1.6~\cite{liu2024improved}) & 7.8B & 47.2 & 18.4 & 56.5 & 62.6 & 44.0 \\ \midrule
\multicolumn{7}{c}{Training-Free MLLMs} \\ \midrule
Qwen3-VL-2B~\cite{bai2025qwen3} + REP & 2.0B & 11.7 & 4.1 & 7.6 & 15.9 & 8.7 \\
\rowcolor{gray!10}
Qwen3-VL-2B~\cite{bai2025qwen3} + \textbf{Ours} & 2.0B & \textbf{30.8} & \textbf{14.8} & \textbf{23.3} & \textbf{39.9} & \textbf{24.9} \\
Phi3.5-V~\cite{abdin2024phi} + REP & 4.2B & 22.2 & 13.9 & 11.8 & 28.1 & 17.1 \\
\rowcolor{gray!10}
Phi3.5-V~\cite{abdin2024phi} + \textbf{Ours} & 4.2B & \textbf{35.3} & \textbf{20.6} & \textbf{36.4} & \textbf{56.6} & \textbf{34.0} \\
LFM2.5-VL~\cite{amini2025lfm2} + REP & 1.6B & 24.8 & \textbf{25.6} & 25.4 & 27.2 &  27.2 \\
\rowcolor{gray!10}
LFM2.5-VL~\cite{amini2025lfm2} + \textbf{Ours} & 1.6B & \textbf{36.8} & 24.4 & \textbf{34.5} & \textbf{48.0} & \textbf{33.8}  \\
Qwen2-VL-2B~\cite{wang2024qwen2} + REP & 2.2B & 39.5 & 19.7 & 38.8 & 44.1 & 34.3 \\
\rowcolor{gray!10}
Qwen2-VL-2B~\cite{wang2024qwen2} + \textbf{Ours} & 2.2B & \textbf{48.6} & \textbf{33.3} & \textbf{44.1} & \textbf{52.6} & \textbf{43.3} \\
Qwen2-VL-7B~\cite{wang2024qwen2} + REP & 8.3B & 16.0 & 3.8 & 19.0 & 24.9 & 14.6 \\
\rowcolor{gray!10}
Qwen2-VL-7B~\cite{wang2024qwen2} + \textbf{Ours} & 8.3B & \textbf{47.7} & \textbf{32.1} & \textbf{43.1} & \textbf{54.8} & \textbf{42.6} \\
\bottomrule
\end{tabular}%
}
\end{table}

Before hard negative mining can be effective, the MLLM's generative space must be aligned for discrimination. Our hierarchical embedding prompt achieves this without parameter updates. As illustrated by the UMAP visualization in Fig. \ref{fig:umap_visualization}, the base model without system-level prompting suffers from a severe modality gap. Applying our hierarchical prompt decisively closes this gap, structurally organizing heterogeneous inputs into semantically coherent clusters. This latent conditioning translates to exceptional zero-shot performance (Table \ref{tab:zeroshot_mmeb}), rivaling models that underwent massive contrastive pre-training. Furthermore, this structural constraint is vital for training stability. As shown in Fig.~\ref{fig:training_dynamics}, fine-tuning without the system prompt leads to representation collapse on out-of-distribution tasks, whereas our hierarchical conditioning ensures monotonic, stable learning curves.

\subsubsection{Effectiveness of Negative Sampling Strategies}

\begin{table}[t]
\centering
\caption{\textbf{Performance comparison of various negative sampling strategies}. The training time is based on 1.5 epochs. $B$ denotes batch size, not effective batch size.}
\label{tab:ablation_sampling}
\resizebox{\columnwidth}{!}{%
\begin{tabular}{@{}llcccccc@{}}
\toprule
\multirow{2}{*}{Pmt.} & \multirow{2}{*}{Sampling} & \multicolumn{2}{c}{Hparams} &\multirow{2}{*}{\begin{tabular}[c]{@{}l@{}} Training \\ Time (hr) \end{tabular}} & \multicolumn{3}{c}{Performance} \\
\cmidrule(lr){3-4} \cmidrule(l){6-8}
 &  & $k$ & $B$ &  & IND & OOD & Avg. \\
\midrule
- & In-batch & - & 1024 & - & 66.0 & 52.6 & 59.3 \\
\midrule
\multirow{8}{*}{Ours} & In-batch & - & 1024 & 13.7 & 68.5 & 58.6 & 64.1 \\
\cmidrule(l){2-8}
& \multirow{2}{*}{HN} & 1 & 256 & 20.7 & 60.8 & 51.4 & 56.6 \\
& & 7 & 256 & 59.7 & 62.6 & 54.0 & 58.8 \\
\cmidrule(l){2-8}
& HN + $\beta$~\cite{gu2025breaking} & 7 & 256 & 59.7 & 67.2 & 58.1 & 63.1 \\
\cmidrule(l){2-8}
& SaHa & 7 & 256 & 15.8 & \textbf{70.8} & 62.0 & 66.9 \\
& \text{  }\textit{w/o} cluster & 7 & 256 & 55.9 & 68.1 & 60.8 & 64.9 \\ \cmidrule(lr){2-8}
& SaHa& 15 & 128 & 16.1 & 70.5 & \textbf{63.4} & \textbf{67.4} \\

\bottomrule
\end{tabular}
}
\vspace{-6pt}
\end{table}

Table~\ref{tab:ablation_sampling} compares SaHa against standard sampling baselines. Conventional hard negative (HN) mining degrades performance because it frequently treats unlabeled positives as negatives. While filtering heuristics like HN+$\beta$~\cite{gu2025breaking} marginally improve this, they remain fragile. SaHa not only achieves the highest overall performance by strictly filtering false negatives, but it also dramatically reduces training time by organizing samples into mutually hard, computationally efficient clusters. To disentangle the contribution of filtering in the query space from the dense supervision provided by mutually hard clusters, we evaluate a variant of SaHa without clustering, denoted as SaHa \textit{w/o} cluster. In this configuration, each anchor is paired with hard negatives ($k=7$) filtered in the query space, and the samples are processed independently without cluster batching. We maintain identical hyperparameters for a fair comparison. Notably, even without clustering, this variant outperforms both standard HN and the existing filtering heuristic (HN$+\beta$). This validates that our query-space filtering is more effective at mitigating false negatives than conventional similarity-based thresholding. Furthermore, the mutually hard cluster design not only dramatically reduces the training time but also yields an additional performance boost. Note that the effective batch size for HN and SaHa is calculated as $(k+1) \times B$. In SaHa, the actual size dynamically adapts and may be smaller for tasks with restricted candidate pools (\textit{e.g.}, Classification) where the number of valid negatives is less than $k$.

\subsubsection{False Negative Mitigation}

\begin{table}[t]
\centering
\caption{\textbf{False Negative (FN) contamination.} (a) Absolute exact FN rates via image-to-image retrieval ($k=16, m=5$). (b) Latent high-risk FN rates (semantic overlap $\ge 0.90$).}
\label{tab:fn_analysis}

\begin{minipage}[t]{0.48\columnwidth}
    \centering
    {\small (a) Absolute FN (\%)} \\[1mm] 
    \resizebox{\linewidth}{!}{%
    \begin{tabular}{@{}lcc@{}}
    \toprule
    \textbf{Task} & \textbf{HNS} & \textbf{SaHa} \\
    \midrule
    ImageNet-1K & 14.52 & \textbf{11.51} \\
    N24News     & 0.05  & \textbf{0.00} \\
    SUN397      & 14.92 & \textbf{11.91} \\ \midrule
    \textit{Average} & \textit{9.83} & \textit{\textbf{7.81}} \\
    \bottomrule
    \end{tabular}%
    }
\end{minipage}\hfill
\begin{minipage}[t]{0.48\columnwidth}
    \centering
    {\small (b) Latent High-Risk FN (\%)} \\[1mm] 
    \resizebox{.89\linewidth}{!}{%
    \begin{tabular}{@{}lcc@{}}
    \toprule
    \textbf{Task} & \textbf{HNS} & \textbf{SaHa} \\
    \midrule
    WebQA   & 15.17 & \textbf{0.00} \\
    CIRR    & 80.03 & \textbf{34.67} \\
    NIGHTS  & 96.19 & \textbf{57.48} \\
    A-OKVQA & 19.49 & \textbf{0.00} \\
    OK-VQA  & 22.72 & \textbf{0.26} \\
    DocVQA  & 24.03 & \textbf{2.34} \\
    ChartQA & 8.17  & \textbf{0.02} \\
    \bottomrule
    \end{tabular}%
    }
\end{minipage}
\vspace{-4pt}
\end{table}

To validate SaHa's core mechanism, we evaluate the false negative (FN) contamination rates under two distinct protocols in Table~\ref{tab:fn_analysis}.

\begin{figure}
\centering
\includegraphics[width = 1.\columnwidth]{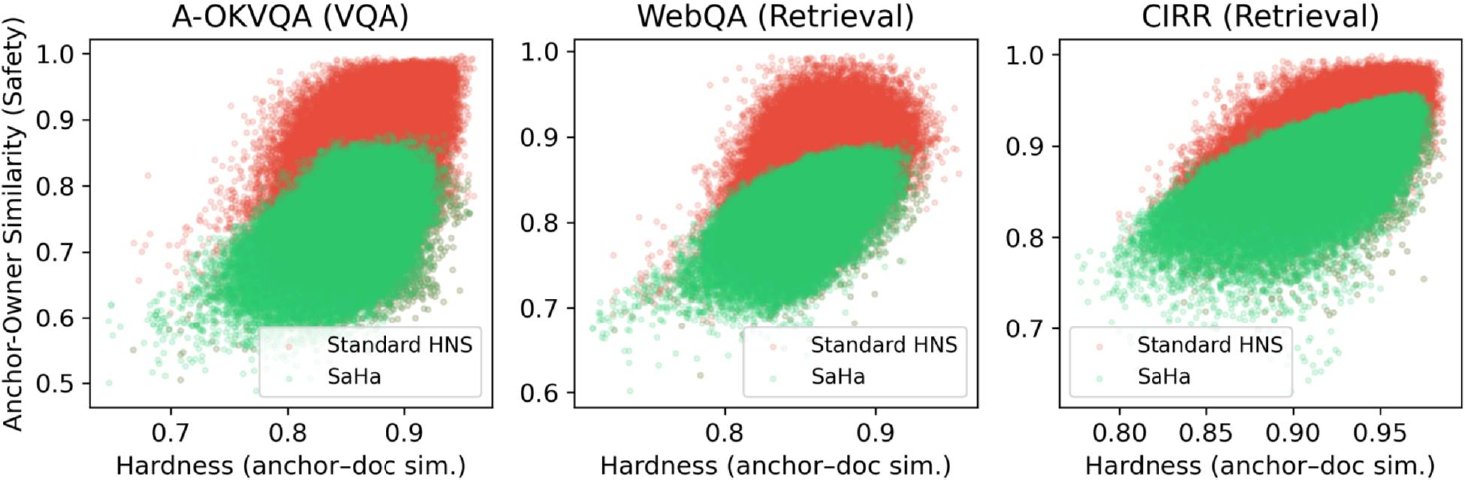}
\captionof{figure}{
    \textbf{Hardness vs. Safety trade-off.} Compared to HNS (red), SaHa (green) effectively lowers semantic overlap (y-axis) while preserving task difficulty (x-axis).}
\vspace{-8pt}
\label{fig:hardness}
\end{figure}

First, to establish an absolute ground truth, we reformulate Classification tasks into an image-to-image retrieval problem. After randomly sub-sampling 10,000 images per dataset, standard HNS still frequently retrieves images of the exact same class. This confirms that blindly maximizing visual similarity inherently leads to semantic collapse. Conversely, SaHa effectively mitigates this contamination by structurally avoiding the extremely high-similarity danger zone. By curating mutually hard negatives from a safer boundary, SaHa consistently reduces the absolute FN rate across all evaluated datasets (reducing the average exact FN rate from 9.83\% to 7.81\%) without requiring any ground-truth label supervision (Table \ref{tab:fn_analysis}a).

Second, for open-ended Retrieval and VQA tasks, we analyze latent false negatives, sampled negatives with dangerously high semantic overlap (anchor-owner similarity $\ge 0.90$). Standard HNS suffers from massive contamination, especially in fine-grained datasets. By mapping candidates back to their origin, SaHa isolates mutually hard negatives while structurally avoiding semantic overlap, drastically reducing the high-risk FN ratio across all evaluated datasets (Table~\ref{tab:fn_analysis}b). Specifically, SaHa eliminates the vast majority of latent FNs in VQA tasks (dropping the average from 18.6\% to 0.7\%) and roughly halves them in retrieval tasks (from 58.6\% to 29.1\%).

To provide a deeper intuitive understanding, Fig. \ref{fig:hardness} visualizes the inherent trade-off between negative hardness (x-axis: visual similarity to the anchor) and safety (y-axis: semantic overlap risk). In this space, an ideal hard negative should be located in the bottom-right corner, confusing (high x-value) but semantically distinct (low y-value). Standard HNS blindly relies on similarity, which inevitably pushes the sampled negatives into the upper-right. This visually confirms that pursuing extreme hardness without semantic checks guarantees severe false negative contamination. In contrast, our SaHa strategy effectively breaks this destructive correlation. By cross-checking candidates against their owner queries, SaHa successfully shifts the sampling distribution downward along the y-axis into the safe zone. Notably, we achieve this while completely preserving the spread along the x-axis, demonstrating its ability to curate challenging negatives. Quantitatively, the reduction in semantic overlap is 2.9$\times$ (Retrieval) and 3.7$\times$ (VQA) larger than the incidental reduction in hardness.

Importantly, this false-negative reduction directly translates to downstream performance gains. We observe a strong positive correlation (Pearson $r \approx +0.55$) between the drop in latent false negatives and the task-specific accuracy improvements. Subsets that suffered from heavy latent contamination under standard HNS (\textit{e.g.}, MSCOCO-I2T, OK-VQA) yielded the largest absolute accuracy gains (+18.5 and +15.3 percentage points, respectively). This confirms that SaHa mitigates the core bottleneck of semantic collapse in multimodal embedding spaces.

\subsubsection{Hyperparameter Sensitivity}

\begin{table}[h!]
\centering
\caption{\textbf{Sensitivity analysis for the pool multiplier in SaHa ($k=7$).} The pool multiplier denotes the size of the initial candidate pool relative to the number of negatives ($k$)}
\label{tab:ablation_multiplier}
\begin{tabular}{l c ccc}
\toprule
\multirow{2}{*}{Metric} & \multirow{2}{*}{Baseline} & \multicolumn{3}{c}{Pool Multiplier} \\
\cmidrule(lr){3-5}
 & & 4 & 6 & 8 \\
\midrule
IND & 62.6 & \textbf{70.8} & 70.7 & 70.6  \\
OND & 54.0 & \textbf{62.0} & 61.9 & 61.6 \\ \midrule
Avg. & 58.8 & \textbf{66.9} & 66.8 & 66.6 \\
\bottomrule
\end{tabular}
\end{table}

We confirm the robustness of our strategy through a sensitivity analysis on the pool multiplier $m$. As shown in Table~\ref{tab:ablation_multiplier}, all tested multiplier settings not only achieve substantial gains over the baseline, but the performance also remains stable, showing only a slight degradation as the multiplier increases. This low sensitivity validates the robustness and practicality of SaHa, since this strategy delivers high performance without requiring meticulous hyperparameter tuning.

\section{Conclusion}
\label{sec:conclusion}

In this paper, we presented a highly efficient framework that adapts generative Multimodal Large Language Models (MLLMs) into universal embedding models without the need for resource-intensive contrastive pre-training. We achieve this by introducing a hierarchical embedding prompt that provides strong latent conditioning, effectively bridging the modality gap and unlocking powerful zero-shot capabilities. Building upon this latent conditioning, we proposed Self-aware Hard Negative Sampling (SaHa), which shifts the negative mining mechanism to the query-space. By tracing candidates back to their owner queries, SaHa rigorously filters out false negatives to prevent semantic collapse, while constructing mutually hard clusters for highly efficient batch training. Extensive experiments demonstrate that our unified approach achieves state-of-the-art performance and training efficiency on the massive MMEB benchmark. Furthermore, successful evaluations on emerging architectures and the unseen video modality confirm the fundamental robustness and cross-modal generalization of our method.

\bibliographystyle{IEEEtran}
\bibliography{references}



\begin{IEEEbiography}[{\includegraphics[width=1in,height=1.25in,clip,keepaspectratio]{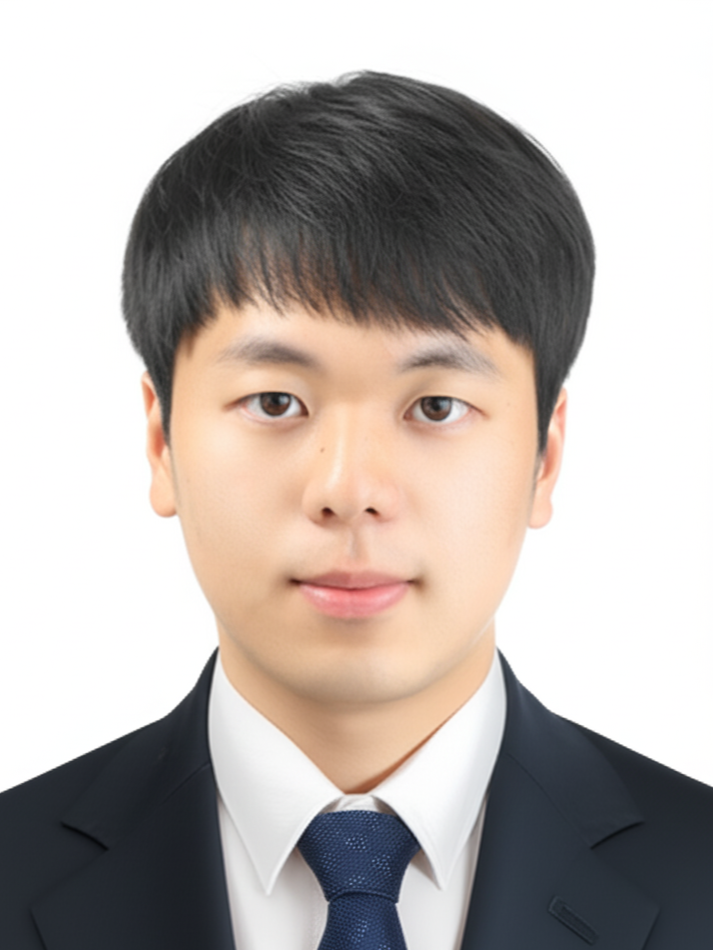}}]{Yeong-Joon Ju} received the B.S. degree in computer engineering from Sejong University, Seoul, Republic of Korea, in 2020. He is currently a Ph.D. student in the Department of Artificial Intelligence at Korea University, Seoul, Republic of Korea. His current research interests include multimodal retrieval, multimodal embedding models, and LLM-based agents.

\end{IEEEbiography}

\begin{IEEEbiography}[{\includegraphics[width=1in,height=1.25in,clip,keepaspectratio]{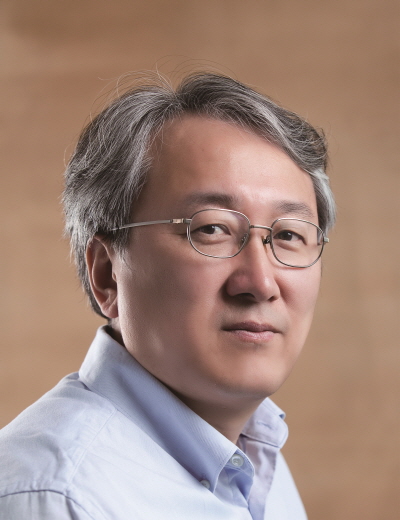}}]{Seong-Whan Lee} (Fellow, IEEE) received the B.S. degree in computer science and statistics from Seoul National University, Seoul, Republic of Korea, in 1984, and the M.S. and Ph.D. degrees in computer science from the Korea Advanced Institute of Science and Technology (KAIST), Seoul, Republic of Korea, in 1986 and 1989, respectively. He is currently the head of the Department of Artificial Intelligence, Korea University, Seoul, Republic of Korea. His current research interests include artificial intelligence, pattern recognition, and brain engineering. Dr. Lee is a fellow of the International Association of Pattern Recognition and the Korea Academy of Science and Technology.
\end{IEEEbiography}

\newpage

\appendix

\begin{algorithm}
    \begin{footnotesize}
    \caption{Self-aware Hard Negative Sampling}
    \label{alg:shns}
    \DontPrintSemicolon
    \SetKwInput{KwIn}{Input}
    \SetKwInput{KwOut}{Output}
    
    \SetKwProg{Fn}{Function}{}{end}

    \KwIn{
        $Q_{full}$: The entire set of training queries\;
        $C$: The entire set of candidates\; 
        $Owner$: A dictionary mapping a candidate to its owner queries\; 
        $k$: The number of hard negatives to sample per cluster\;
        $m$: Pool multiplier for the initial candidate pool size.
    }
    \KwOut{
        $H$: The final training set of mutually hard clusters.
    }
    
    \BlankLine
    
    \Fn{\text{ConstructClusters}($Q_{sub}, U_{global}, \text{allow\_redundancy}$)}{
        $H_{local} \leftarrow \emptyset$\;
        $U_{local} \leftarrow \emptyset$\;
    
        \For{each anchor query $q \in Q_{sub}$}{
            \If{$q \notin U_{global}$}{
                \tcp{Step 1: Candidate Mining}
                $D_{cand} \leftarrow \text{Retrieve}(q, I_C)[:m\times k]$\;
                $Q_{owners} \leftarrow \emptyset$\; 
                
                \tcp{Step 2: Owner Query Identification}
                \For{each candidate $c \in D_{cand}$}{ 
                    \If{$c \in Owner$}{
                        \tcp{Find the owner query most similar to the anchor q}
                        $q_{owner} \leftarrow \text{Retrieve}(q, Owner[c])$\; 
                        $Q_{owners} \leftarrow Q_{owners} \cup \{q_{owner}\}$\; 
                    }
                }
                
                \lIf{\text{allow\_redundancy}}{$Q_{owners} \leftarrow Q_{owners} \setminus U_{local}$}
                \lElse{$Q_{owners} \leftarrow Q_{owners} \setminus U_{global}$}
                
                \tcp{Step 3: Self-aware Filtering}
                $Q_{neg} \leftarrow \text{SortByAscendingSimilarity}(Q_{owners}, q)[:k]$\; 
                
                \tcp{Construct a mutually hard cluster of size k+1}
                $C_{cluster} \leftarrow \{(q, c^+)\} \cup \{(q_j, c_j^+) \mid q_j \in Q_{neg}\}$\;
                $H_{local} \leftarrow H_{local} \cup \{C_{cluster}\}$\;
                
                $U_{global} \leftarrow U_{global} \cup \{q\} \cup Q_{neg}$\;
                \lIf{\text{allow\_redundancy}}{$U_{local} \leftarrow U_{local} \cup Q_{neg}$}
            }
        }
        \KwRet{$H_{local}$}\;
    }
    
    \BlankLine
    
    \tcp{Main Process}
    $H \leftarrow \emptyset$, $U \leftarrow \emptyset$\;
    $I_C \leftarrow \text{BuildGlobalIndex}(C)$\;
    
    \nl \textbf{Phase 1: Primary Mining (Strictly Non-overlapping)}\;
    $H \leftarrow \text{ConstructClusters}(Q_{full}, U, \text{False})$\;
    
    \BlankLine
    \nl \textbf{Phase 2: Leftover Processing (Allow Limited Redundancy)}\;
    $Q_{leftover} \leftarrow Q_{full} \setminus U$\;
    $H \leftarrow H \cup \text{ConstructClusters}(Q_{leftover}, U, \text{True})$\;
    
    \KwRet{$H$}\;
    \end{footnotesize}
\end{algorithm}

\subsection{Algorithm Details: Self-aware Hard Negative Sampling}
To ensure reproducibility, we provide the detailed procedure of the proposed Self-aware Hard Negative Sampling (SaHa) in Algorithm \ref{alg:shns}. The algorithm leverages a query-space filtering mechanism to strictly eliminate potential false negatives while constructing mutually hard clusters for efficient training.

\subsection{Hierarchical Prompt Templates}
We apply our hierarchical embedding prompts to map inputs into a shared generative latent space. The exact templates used for queries and candidate documents are as follows:

\begin{tcolorbox}[colback=gray!5!white,colframe=gray!75!black,title=Prompt format for queries]
    \textbf{System:} Given an image, summarize the provided image in one word. Given only text, describe the text in one word.
    
    \textbf{User:} $\lbrack$ Task Instruction  $\rbrack$
    
    $\lbrack$ Query $\rbrack$ $\lbrack$ Representation Prompt $\rbrack$
    
    \textbf{Assistant:}
\end{tcolorbox}

\begin{tcolorbox}[colback=gray!5!white,colframe=gray!75!black,title=Prompt format for candidates]
    \textbf{System:} Given an image, summarize the provided image in one word. Given only text, describe the text in one word.

    \textbf{User:} $\lbrack$ Candidate $\rbrack$
    
    \textbf{Assistant:}
\end{tcolorbox}

Here, we simply set the \texttt{[Representation Prompt]} as: \textit{``Represent the given text/image in one word."}

\subsection{Dataset Statistics and Descriptions}

\begin{table}[]
\centering
\caption{\textbf{The statistics of the MMEB dataset.}}
\label{tab:mmeb_stat}
\resizebox{.99\columnwidth}{!}{
\begin{tabular}{llrr}
\toprule
\textbf{Meta-Task} & \textbf{Dataset} & \textbf{\#Train} & \textbf{\#Eval} \\
\midrule
\multirow{10}{*}{\begin{tabular}[c]{@{}l@{}}Classification\\ (10 Tasks)\end{tabular}} & ImageNet-1K & 100K & 1000 \\
 & N24News & 49K & 1000 \\
 & HatefulMemes & 8K & 1000 \\
 & VOC2007 & 8K & 1000 \\
 & SUN397 & 20K & 1000 \\
 & Place365 & - & 1000 \\
 & ImageNet-A & - & 1000 \\
 & ImageNet-R & - & 1000 \\
 & ObjectNet & - & 1000 \\
 & Country-211 & - & 1000 \\
\midrule
\multirow{10}{*}{\begin{tabular}[c]{@{}l@{}}VQA\\ (10 Tasks)\end{tabular}} & OK-VQA & 9K & 1000 \\
 & A-OKVQA & 17K & 1000 \\
 & DocVQA & 40K & 1000 \\
 & InfographicVQA & 24K & 1000 \\
 & ChartQA & 28K & 1000 \\
 & Visual7W & 70K & 1000 \\
 & ScienceQA & - & 1000 \\
 & VizWiz & - & 1000 \\
 & GQA & - & 1000 \\
 & TextVQA & - & 1000 \\
\midrule
\multirow{12}{*}{\begin{tabular}[c]{@{}l@{}}Retrieval\\ (12 Tasks)\end{tabular}} & VisDial & 123K & 1000 \\
 & CIRR & 26K & 1000 \\
 & VisualNews\_t2i & 100K & 1000 \\
 & VisualNews\_i2t & 100K & 1000 \\
 & MSCOCO\_t2i & 100K & 1000 \\
 & MSCOCO\_i2t & 113K & 1000 \\
 & NIGHTS & 16K & 1000 \\
 & WebQA & 17K & 1000 \\
 & OVEN & - & 1000 \\
 & FashionIQ & - & 1000 \\
 & EDIS & - & 1000 \\
 & Wiki-SS-NQ & - & 1000 \\
\midrule
\multirow{4}{*}{\begin{tabular}[c]{@{}l@{}}Visual Grounding\\ (4 Tasks)\end{tabular}} & MSCOCO & 100K & 1000 \\
 & Visual7W-Pointing & - & 1000 \\
 & RefCOCO & - & 1000 \\
 & RefCOCO-Matching & - & 1000 \\
\bottomrule
\end{tabular}
}
\end{table}

\subsubsection{MMEB Dataset Distribution}
The Massive Multimodal Embedding Benchmark (MMEB) comprises 36 diverse datasets spanning four meta-tasks. Table \ref{tab:mmeb_stat} details the distribution of training and evaluation samples across the datasets used in our experiments.

\subsubsection{Fine-grained Compositionality Benchmarks}
To deeply analyze the model's fine-grained visual-linguistic alignment, we employed the SugarCrepe and SugarCrepe++ benchmarks. Unlike standard retrieval, these datasets evaluate sensitivity to subtle semantic and lexical alterations. SugarCrepe consists of approximately 10.5K evaluation pairs, generated to test the model against negative captions with swapped objects, swapped attributes, or replaced relations. SugarCrepe++, an advanced extension, comprises meticulously curated subsets specifically for testing Image-to-Text (ITT) and Text-only (TOT) retrieval capabilities to evaluate fundamental compositional understanding.

\begin{figure*}[htbp]
    \centering
    \includegraphics[width=.93\textwidth]{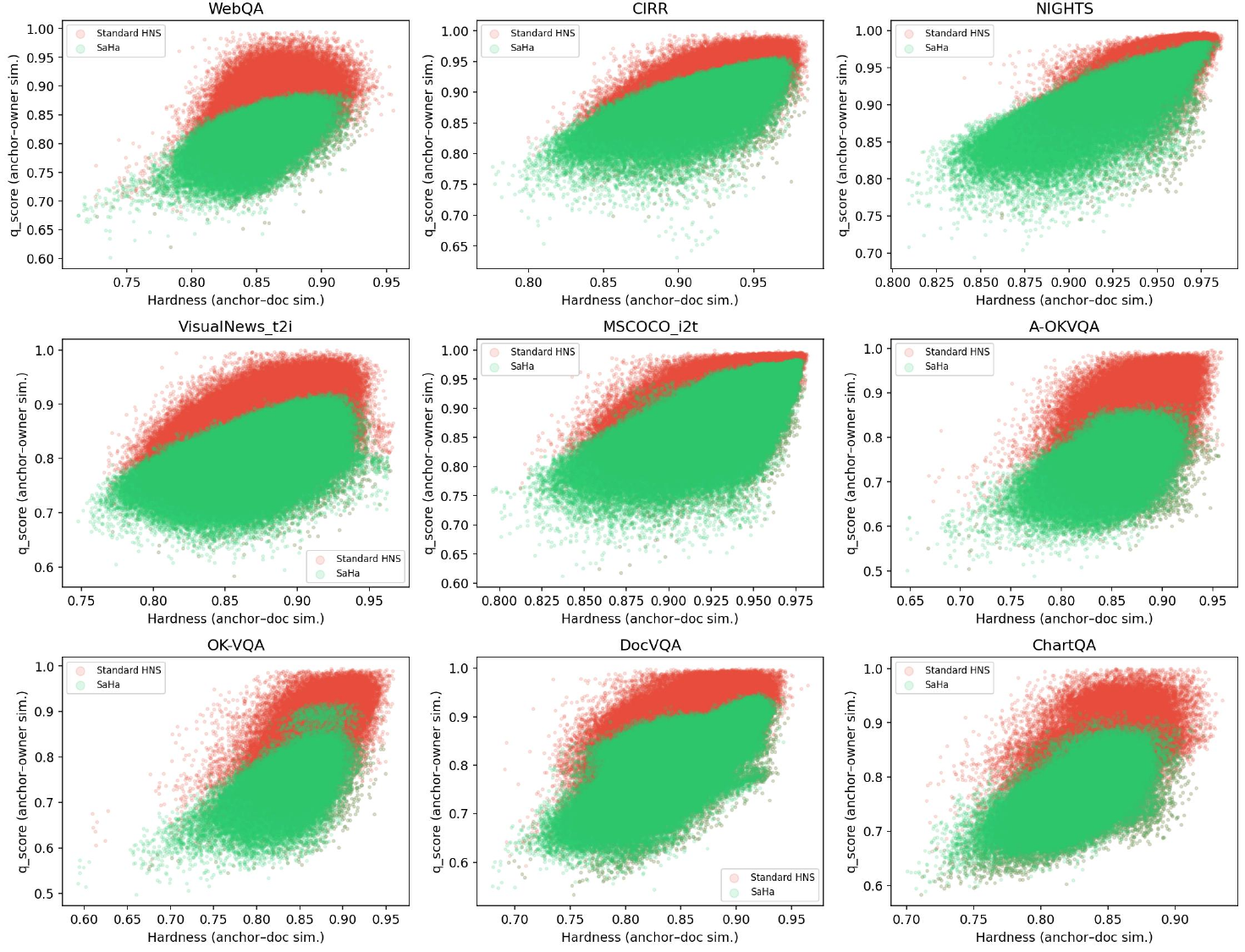}
    \caption{\textbf{Comprehensive Hardness vs. Safety trade-off across 9 datasets in MMEB.} SaHa consistently mitigates false negative contamination while preserving candidate-space hardness across various domains. \texttt{q\_score} denotes anchor-owner similarity. }
    \label{fig:appendix_scatter}
\end{figure*}

\subsubsection{Video Modality Tasks (MMEBv2 Protocol)}

To explicitly assess cross-modality generalization (Section IV.B.3), we evaluated our model on six diverse video tasks. The details of these datasets are as follows:

\begin{itemize}
    \item \textbf{ActivityNetQA}: Focuses on open-ended video question answering within the human activity domain. It requires the model to process both video and text to generate textual answers. In our evaluation, we utilize 1,000 queries, each paired with 2 candidates.
    \item \textbf{Breakfast}: Contains around 1.9K crowdsource video clips in the wild, totaling more than 70 hours, concerning the preparation of 10 different types of breakfast (e.g., cereal, milk, pancakes). We used all 433 sample clips filmed with camera 01. The candidate texts are the 10 types of breakfast.
    \item \textbf{EgoSchema}: A diagnostic benchmark for long-form video understanding, comprising over 5,000 multiple-choice QA pairs spanning more than 250 hours of egocentric video. Each question targets long-range temporal reasoning. We use a subset of 500 questions with publicly available annotations.
    \item \textbf{Kinetics-700-2020}: Made up of approximately 648K YouTube video clips covering around 700 human action labels. We sampled 1K video-answer pairs from the validation set, with candidate texts being the list of all labels.
    \item \textbf{UCF101}: An open-domain video dataset consisting of approximately 13K videos across 101 action categories. We sampled 1K clip-text pairs from the test splits, with candidate texts being all 101 action categories.
    \item \textbf{Video-MME}: A full-spectrum benchmark for evaluating MLLMs on video understanding. It consists of 2,700 manually annotated QA pairs based on 900 videos. It ensures broad scenario coverage and captures diverse temporal dynamics. We evaluated on 1,000 queries.
\end{itemize}

\begin{figure*}
\centering
\includegraphics[width = 1.\textwidth]{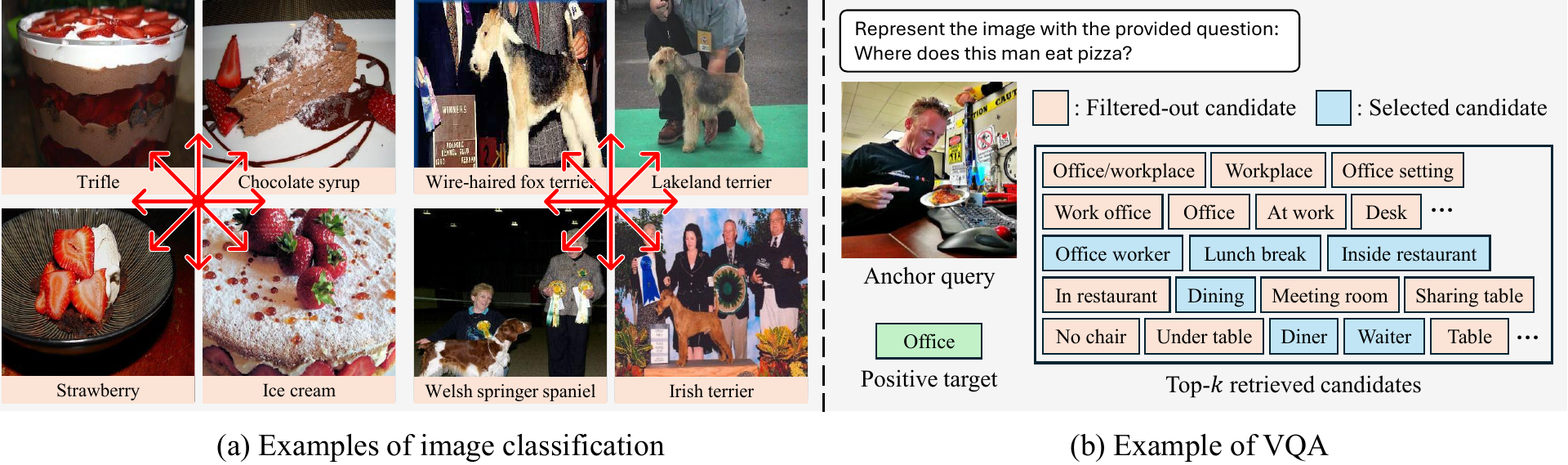}
\captionof{figure}{
    \textbf{Qualitative examples illustrating SaHa's sampling mechanism.} (a) SaHa groups mutually hard negative images within the same thematic cluster. (b) SaHa effectively filters out potential false negatives (pink boxes) that severely overlap with the positive target (green box) and selects valid hard negatives (blue boxes).}
\label{fig:shns_examples}
\vspace{-8pt}
\end{figure*}

\subsection{Comprehensive Visualizations of Negative Sampling}
Due to space constraints in the main text, we present the comprehensive Hardness vs. Safety trade-off visualizations across 9 evaluated fine-grained retrieval and VQA datasets in Fig.~\ref{fig:appendix_scatter}. Consistent with our analysis in Section IV.C, standard Hard Negative Sampling (HNS, red) aggressively samples candidates with dangerously high semantic overlap. In contrast, our SaHa strategy (green) effectively pulls the distribution downward into the safe zone while fully preserving the task difficulty (hardness) across all modalities and domains.

Furthermore, to provide a more intuitive understanding of this mechanism, Fig.~\ref{fig:shns_examples} illustrates qualitative examples of the sampled negatives. While standard HNS often blindly retrieves top-ranked candidates that are semantically identical to the positive target, SaHa successfully navigates this challenge. As shown in Fig.~\ref{fig:shns_examples}a, SaHa groups mutually hard negatives from the same thematic cluster, such as desserts containing strawberries, or dog breeds with similar profiles, but strictly separates their exact class identities. In VQA tasks (Fig.~\ref{fig:shns_examples}b), although perfectly isolating semantic overlap in open-ended textual candidates is inherently difficult, SaHa effectively identifies and discards high-risk false negatives (\textit{e.g.}, ``Workplace", ``At work"). Guided by the structural constraints of the owner queries, SaHa reliably curates true hard negatives (\textit{e.g.}, ``Office worker", ``Lunch break") that share the overarching visual or thematic context but remain semantically distinct from the exact answer.

\subsection{Qualitative Analysis of Failure Cases}

\begin{figure*}
    \centering
    \includegraphics[width=1.\textwidth]{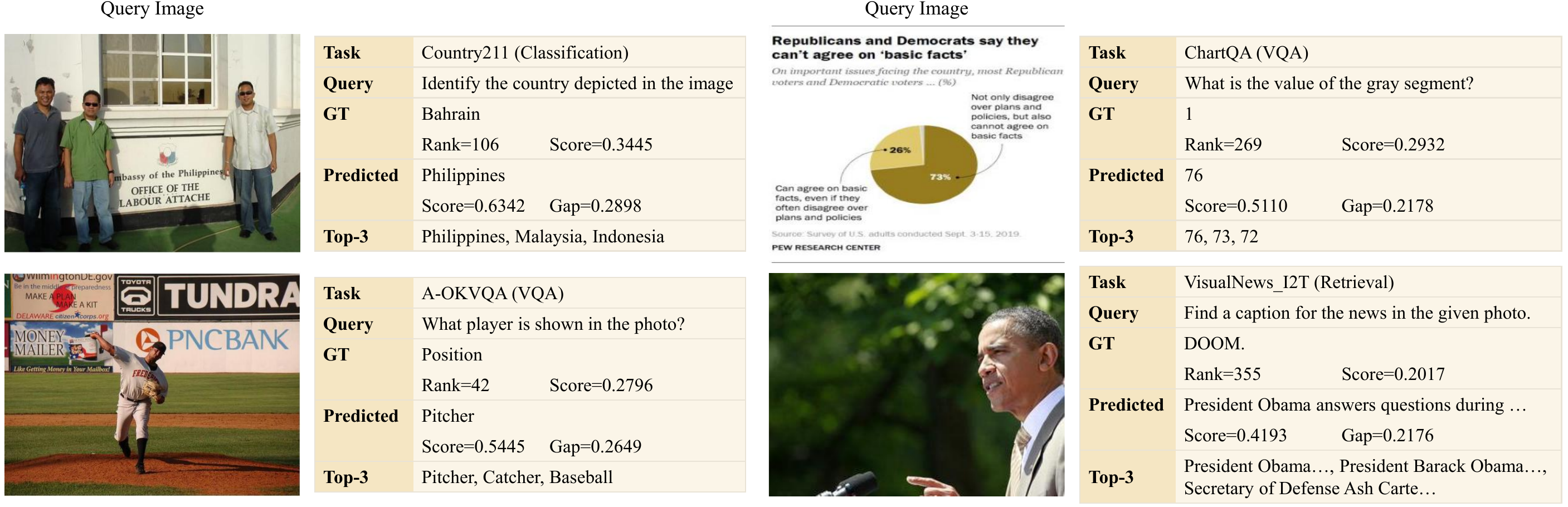}
    \caption{\textbf{Qualitative analysis of retrieval failure cases.} Failures typically occur due to (1) an over-reliance on dominant OCR cues without deeper contextual reasoning (top-left), (2) the absence of multi-step mathematical deduction (top-right), (3) ambiguous or counter-intuitive ground-truth annotations (bottom-left), and (4) the semantic gap between literal visual descriptions and highly abstract news headlines (bottom-right).}
    \label{fig:failure_cases}
\end{figure*}

While our framework demonstrates robust universal embedding capabilities across diverse domains, analyzing its failure cases provides valuable insights into the boundaries of the current representation paradigm. Fig.~\ref{fig:failure_cases} illustrates four representative failure modes encountered during evaluation.

We observe that failures primarily stem from the extreme compression required by the ``in one word" representation, which struggles with tasks demanding complex, multi-step reasoning. For instance, in the \textbf{Country211} task (top-left), the model correctly extracts the OCR text ``Philippines" from the sign but fails to deduce the contextual reasoning that an embassy is typically located in a foreign country (Bahrain). Similarly, in the \textbf{ChartQA} example (top-right), the model fails to perform the implicit mathematical deduction ($100\% - 26\% - 73\% = 1\%$) required to determine the value of the unlabeled gray segment. 

Furthermore, some failures are inherently tied to dataset artifacts or the semantic gap between modalities. In the \textbf{A-OKVQA} example (bottom-left), the model visually comprehends the image and accurately predicts ``Pitcher". However, it is penalized because the ground truth is broadly and somewhat counter-intuitively annotated as ``Position". Lastly, mapping literal visual features to highly abstract textual queries remains challenging. In the \textbf{VisualNews} example (bottom-right), the image of Barack Obama is paired with the highly abstract ground-truth caption ``DOOM." The model naturally retrieves a visually consistent and descriptive caption, which is considered incorrect under the strict exact-match evaluation protocol. These cases highlight the necessity for future work to explore multi-vector representations or adaptive reasoning modules to bridge these remaining gaps.

\subsection{Granular Per-Task Performance Analysis}

\begin{figure}
    \centering
    \includegraphics[width=1.02\columnwidth]{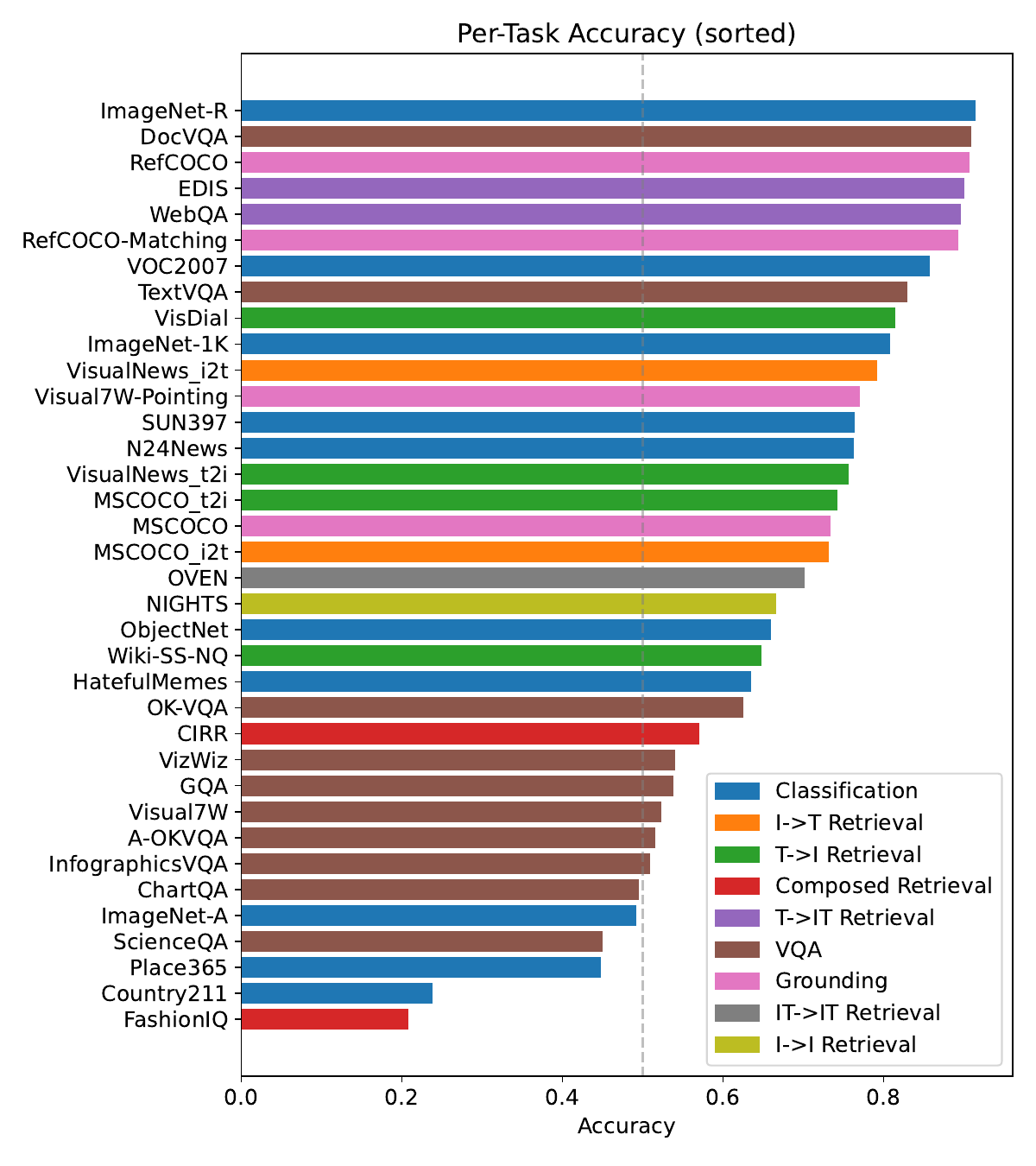}
    \caption{\textbf{Per-task accuracy distribution across the 36 MMEB datasets.} The performance of our SaHa-trained Qwen2-VL-2B model is sorted in descending order. Colors denote the specific meta-task categories.}
    \label{fig:per_task_acc}
\end{figure}

To provide a more comprehensive understanding of our model's capabilities across diverse domains, Fig.~\ref{fig:per_task_acc} illustrates the detailed per-task accuracy of the Qwen2-VL-2B model trained with our SaHa framework. The 36 datasets from the MMEB benchmark are sorted in descending order of accuracy and color-coded by their respective meta-task categories.

We observe that Visual Grounding (pink) and various Image-Text Retrieval tasks generally achieve high performance. Interestingly, a significant performance variance is observed within the VQA category (brown). Tasks such as DocVQA and TextVQA achieve exceptionally high accuracy, ranking near the top. This is primarily because these datasets often require entity-centric responses (\textit{e.g.}, exact OCR string matching), where the positive target is lexically and visually highly distinct from other candidates.

Conversely, other open-ended VQA tasks (\textit{e.g.}, ScienceQA, ChartQA, InfographicVQA) cluster toward the middle or lower end of the spectrum. In these scenarios, the candidate pools are often populated with options that share extremely similar semantic contexts or thematic backgrounds. Distinguishing the exact ground truth among such semantically intertwined candidates requires fine-grained, multi-hop reasoning. Compressing this complex reasoning process into a single latent vector makes the discrimination inherently more difficult, highlighting a key challenge and a direction for future research in universal multimodal embeddings.


\vfill

\end{document}